\documentclass[11pt]{article}

\usepackage[final]{acl}

\usepackage{booktabs} 
\usepackage{multirow}
\usepackage{amsmath}
\usepackage{amsthm}
\usepackage{amssymb}
\usepackage{hyperref}
\usepackage[capitalize,noabbrev]{cleveref}
\usepackage{algorithm}
\usepackage{algorithmic}

\usepackage{times}
\usepackage{latexsym}

\usepackage[T1]{fontenc}

\usepackage[utf8]{inputenc}

\usepackage{microtype}

\usepackage{inconsolata}

\usepackage{graphicx}

\newtheorem{assumption}{Assumption}[section]    
\newtheorem{proposition}{Proposition}[section]
\newtheorem{lemma}{Lemma}[section]

\newtheorem{remark}{Remark} 
\title{GeoGR²: Zero-Shot Geospatial Inference via Geostatistically-Guided Iterative Refinement with LLMs}

\author{
Jinfan Tang\textsuperscript{1}\footnotemark[1],
Kunming Wu\textsuperscript{1}\footnotemark[1],
Xieruifeng Gong\textsuperscript{2}\footnotemark[1],
Yuya He\textsuperscript{1} \\
\textbf{HuJie\textsuperscript{1}}, 
\textbf{Wu Junhui\textsuperscript{3}},
\textbf{Yuankai Wu\textsuperscript{1}\footnotemark[2] }\\[1ex]
\textsuperscript{1}College of Computer Science, Sichuan University \\
\textsuperscript{2}National Key Laboratory of Fundamental Science on Synthetic Vision, Sichuan University
\\
\textsuperscript{3}College of Life Science, Sichuan University
}

\begin{document}
\maketitle
\footnotetext[1]{Equal contribution.}
\footnotetext[2]{Corresponding author: \texttt{wuyk0@scu.edu.cn}}
\begin{abstract}
Standard large language model prompting treats geospatial inference as independent, instance-wise prediction, ignoring the fundamental spatial dependencies that govern geographic reality. Consequently, even advanced models struggle with spatial consistency and exhibit severe biases toward populous regions. To bridge this gap, we propose \textbf{GeoGR\textsuperscript{2}} (\textbf{Geo}spatial \textbf{G}raph \textbf{R}efine \textbf{R}easoning), a framework that formalizes zero-shot geospatial prediction as an iterative message-passing process on a dynamically constructed graph. Unlike static retrieval methods, GeoGR\textsuperscript{2} instantiates three dynamic operators via collaborating operators: (1) a \textit{Topology Operator} that constructs graph topology to enforce the Spatial Markov property; (2) a \textit{Feature Operator} that enriches nodes with task-relevant semantic covariates; and (3) an \textit{Update Operator} that performs natural language message passing to iteratively minimize spatial discrepancy. Theoretically, we frame this refinement as a contraction mapping that approximates the fixed point of a global consistency equation. 
Empirically, we validate GeoGR\textsuperscript{2} on diverse physical and socioeconomic tasks. Results demonstrate that by explicitly embedding geostatistical inductive biases, GeoGR\textsuperscript{2} significantly outperforms standard prompting baselines, while effectively mitigating systematic geographic bias. Our framework leverages large language models' intrinsic capacity for understanding spatial correlations through explicit topological scaffolding, without resorting to general graph reasoning paradigms.
\end{abstract}

\section{Introduction}

\begin{figure}[htb!]
    \centering
    \includegraphics[width=\columnwidth]{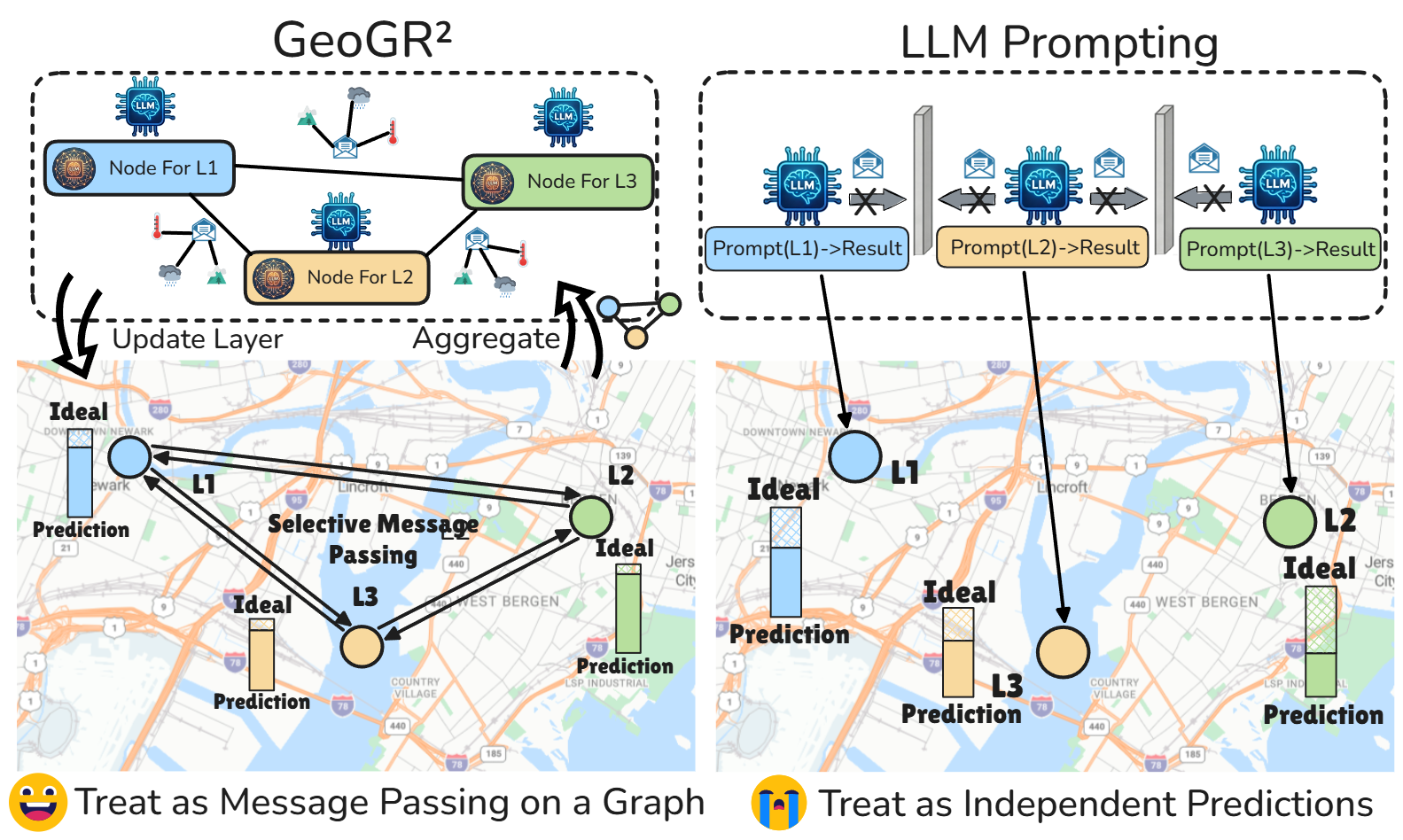} 
    \caption{Conceptual Comparison Between GeoGR\textsuperscript{2}(\textbf{Left}) and LLM Direct Prediction(\textbf{Right}) in geospatial prediction.}
    \label{fig1}
\end{figure}

Large Language Models (LLMs) have emerged as powerful reasoning engines across domains~\cite{li2022competition, boiko2023autonomous}, with growing applications in geospatial intelligence~\cite{manvi2024geollm, li2024urbangpt}. However, a fundamental mismatch persists: \textbf{LLMs process instances independently (i.i.d.) (\cref{fig1}-right), whereas geospatial data is structured by \textit{Tobler’s First Law}~\cite{tobler1970computer}.} Unlike classical geostatistics (e.g., Kriging~\cite{cressie1993statistics}) that explicitly models spatial autocorrelation, standard prompting ignores latent geographic connectivity. Consequently, even advanced models exhibit spatial inconsistency, multi-hop hallucination, and bias toward populous regions due to skewed training distributions~\cite{manvi2024large}.

While Graph Neural Networks (GNNs)~\cite{kipf2016semi, wu2020comprehensive} effectively encode spatial structure via message passing, they require fixed topologies and heavy supervised training~\cite{wu2021inductive}, lacking the semantic flexibility of foundation models. Conversely, current LLM-based geospatial methods rely on costly fine-tuning~\cite{deng2024k2}, which is static and fails to adapt to novel spatial distributions at inference.

Recent work indicates LLMs can grasp spatial dependencies when given explicit topological context~\cite{shang2025survey}. This motivates our hypothesis: LLMs’ latent spatial understanding can be operationalized for geostatistics by \textit{dynamically constructing interaction graphs during reasoning}, transforming standard i.i.d. prompting into a spatially coherent inference paradigm.

To this end, we propose \textbf{GeoGR\textsuperscript{2}} (\textbf{Geo}spatial \textbf{G}raph \textbf{R}efine \textbf{R}easoning), a training-free framework that embeds geostatistical principles directly into the LLM inference loop. Rather than a general graph engine, GeoGR\textsuperscript{2} uses graph structures as a scaffold to operationalize Tobler’s Law. We formulate zero-shot prediction as message passing on a dynamic graph (\cref{fig1}-left), approximating GNN workflows via three natural-language operators: (1) \textbf{Topology Operator} identifies the target’s ``Markov Blanket''; (2) \textbf{Feature Operator} injects semantic covariates; and (3) \textbf{Update Operator} iteratively aggregates neighborhood signals to minimize spatial discrepancy. This loop enables ``spatial analogy'' reasoning, refining predictions through local consensus. Evaluated on physical (Temperature, Precipitation) and socioeconomic (GDP, Mortality) tasks, GeoGR\textsuperscript{2} achieves SOTA zero-shot accuracy while substantially reducing geographic bias. Our contributions are:
\begin{itemize}
    \item \textbf{Conceptual Formulation:} We formalize zero-shot geospatial LLM inference as iterative spatial refinement, bridging classical geostatistics and prompting paradigms.
    \item \textbf{GeoGR\textsuperscript{2} Framework:} A novel training-free architecture that dynamically constructs interaction graphs and performs natural-language neighborhood aggregation, enforcing spatial consistency without fine-tuning.
    \item \textbf{Experimental Validation:} Extensive experiments show consistent superiority over prompting and ML baselines. Theoretically, we show the refinement process acts as a contraction mapping, ensuring convergence to accurate and fair spatial representations.
\end{itemize}

\section{Related Work}

\subsection{LLMs for Geospatial Intelligence}
Geospatial LLMs have evolved from basic prompting or RAG-based coordinate memorization~\cite{manvi2024geollm,zhang2023geogpt} to domain-adaptive pretraining and instruction tuning~\cite{li2024urbangpt,deng2024k2}. While performant, these supervised paradigms demand extensive labeled data and remain sensitive to distribution shifts. Notably, \citet{gurnee2023language} demonstrate that LLMs implicitly encode linear spatiotemporal representations. Instead of costly parameter updates, we \textit{explicitly} activate these latent structures via inference-time graph construction. Although recent evidence confirms LLMs can grasp spatial and relational patterns through explicit topological prompting~\cite{shang2025survey}, this capability remains underexploited in geospatial prediction—a gap our approach directly bridges.

\subsection{Agentic Frameworks and Self-Refinement}
To address the limitations of single-turn prompting, agentic frameworks such as Chain-of-Thought~\cite{wei2022} and ReAct~\cite{yao2023react} facilitate task decomposition, while self-refinement paradigms like Reflexion~\cite{shinn2023} and Self-Refine~\cite{madaan2023} incorporate iterative self-critique. Despite their efficacy in logic and coding domains, these generic strategies lack \textit{domain-specific inductive biases}, typically relying on internal consistency or scalar feedback rather than structured external constraints.

\section{Problem Formulation}

We frame the problem of geospatial inference as a zero-shot regression task over a set of distributed locations. Let $\mathcal{L} = \{\ell_1, \ell_2, \ldots, \ell_n\}$ denote a set of geographic locations, where each $\ell_t$ is defined by its coordinates (latitude, longitude). Associated with each location is a ground-truth scalar variable $y_t \in \mathbb{R}$.

\paragraph{Task Definition.} 
Our objective is to estimate $y_t$ using a pre-trained Large Language Model (LLM) $f_\theta(\cdot)$, without any parameter updates or supervised examples (zero-shot). The estimation is formalized as:
\begin{equation}
    \hat{y}_t = f_\theta(\texttt{prompt}(\ell_t, \mathcal{T})),
\end{equation}
where $\mathcal{T}$ represents the task description. 

\paragraph{Evaluation Metrics.}
Following \citet{manvi2024large}, we assess performance based on two axes: \textit{predictive accuracy} and \textit{fairness}.

\textit{1) Ranking Accuracy ($\rho$):} Since LLMs often output values on arbitrary scales unaligned with real-world units, we measure the structural alignment using Spearman’s rank correlation coefficient $\rho$ between the predicted vector $\hat{\mathbf{y}}$ and ground truth $\mathbf{y}$.

\textit{2) Geographic Bias ($\mathrm{Bias}_y$):} To quantify systematic inequity, we adopt the composite bias metric proposed by \citet{manvi2024large}. This metric penalizes models that exhibit high correlation with a demographic anchor distribution (e.g., population density) while accounting for prediction variance and refusal rates:
\begin{equation}
    \mathrm{Bias}_y(\hat{\mathbf{y}}) = \rho(\hat{\mathbf{y}}, \mathbf{d}) \cdot \mathrm{MAD}(\hat{\mathbf{y}}) \cdot a^2,
    \label{eq:bias}
\end{equation}
where $\mathbf{d} \in \mathbb{R}^n$ is the anchor distribution vector (e.g., population density per location), $\mathrm{MAD}(\cdot)$ denotes the Mean Absolute Deviation from the mean, and $a \in [0, 1]$ is the answer rate. Intuitively, a fair model should yield predictions uncorrelated with the anchor variable $\mathbf{d}$ rather than hallucinating based on population priors.

\section{Method}
\label{sec:method}

\begin{figure*}[t]
    \centering
    \includegraphics[width=0.9\textwidth]{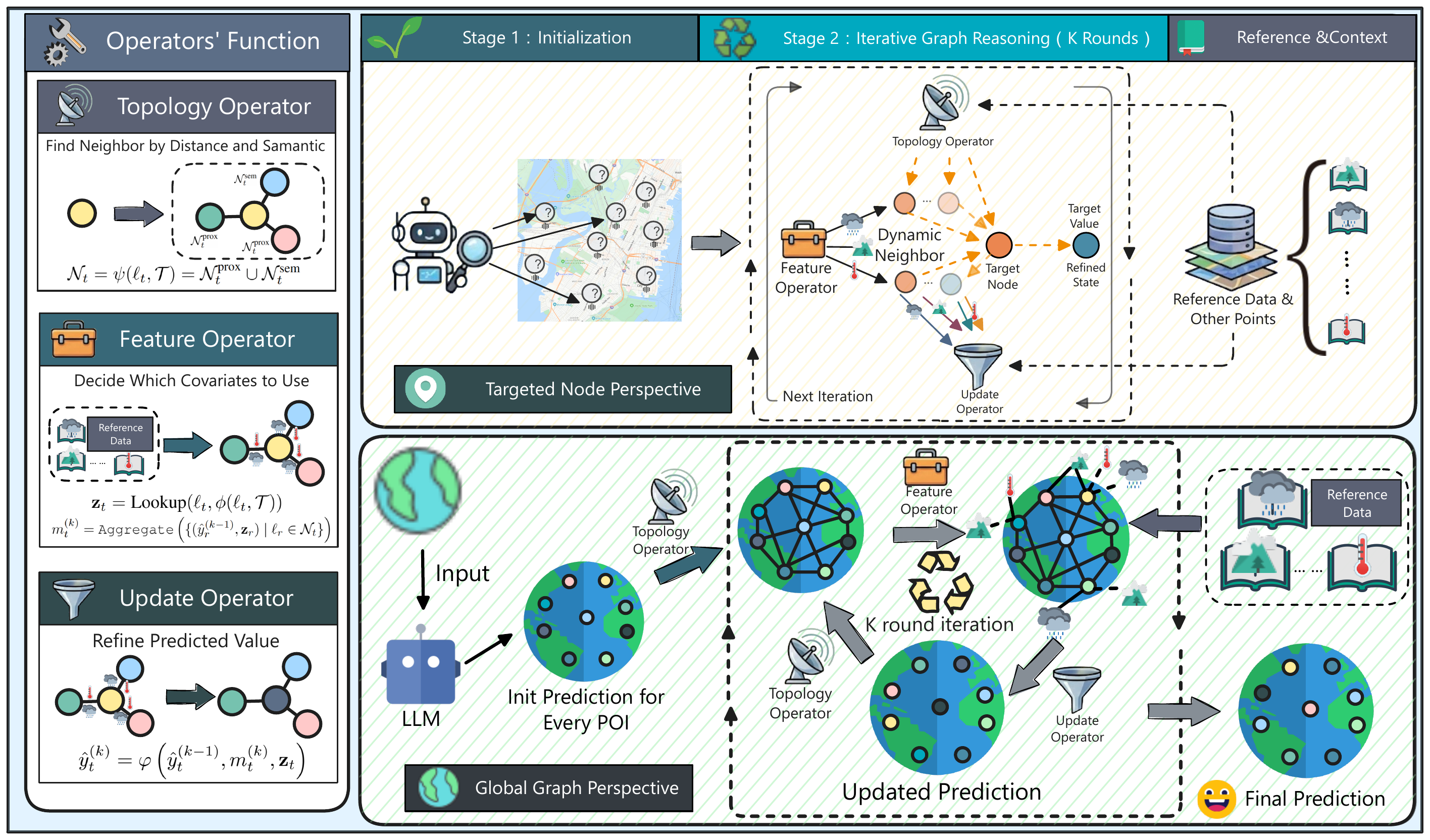} 
    \caption{Overview of GeoGR\textsuperscript{2}. We formulate geospatial prediction as an iterative graph reasoning process where operators dynamically construct spatial topology, select semantic features, and perform natural language message passing to refine predictions.}
    \label{fig2}
\end{figure*}

As illustrated in \cref{fig2}, GeoGR\textsuperscript{2} departs from independent prompting by formalizing inference as an iterative message-passing process driven by three collaborating operators: (1) a \textbf{Topology Operator} that dynamically defines local connectivity; (2) a \textbf{Feature Operator} that enriches node representations; and (3) an \textbf{Update Operator} that performs iterative state refinement.

\subsection{Spatial Dependency Formulation via Graph Structures}
\label{sec:theory}

Our framework is grounded in Probabilistic Graphical Models (PGMs)~\cite{koller2009probabilistic}. We postulate that the joint distribution of geospatial variables follows a Markov Random Field (MRF) structure~\cite{besag1974spatial} regarding an underlying latent graph $\mathcal{G}=(\mathcal{V}, \mathcal{E})$. Here, the graph formalism serves exclusively as a mathematical abstraction for spatial autocorrelation; it is not our objective to advance general-purpose LLM graph reasoning.

To bridge the gap between this abstract formulation and our implementation, we formally define the physical interpretations of the mathematical components:
\begin{itemize}
    \item \textbf{Nodes ($\mathcal{V}$):} The set of discrete geographic locations $\mathcal{L}$.
    \item \textbf{Graph Topology ($\mathbf{W}$):} The binary adjacency matrix $\mathbf{W} \in \mathbb{R}^{n \times n}$ constructed by the \textbf{Topology Operator} ($\psi$). $W_{ij} = 1$ implies location $j$ is selected as a semantic or physical neighbor of $i$.
    \item \textbf{Graph Laplacian ($\mathbf{L}$):} Defined as $\mathbf{L} = \mathbf{D} - \mathbf{W}$ (where $\mathbf{D}$ is the degree matrix). 
    \item \textbf{Refinement Step ($\lambda$):} The scalar $\lambda$ controls the rate of information diffusion during message passing, corresponding to the \textit{trust level} the Update Operator places in neighborhood consensus.
\end{itemize}

Based on the definitions above, we provide three theoretical pillars that justify the design of our framework.

\begin{assumption}[Spatial Markov Property]
\label{asm:markov}
Following \cite{besag1974spatial}, for any location $\ell_t$, given its set of neighbors $\mathcal{N}_t$ (the Markov blanket), the variable $y_t$ is conditionally independent of all other variables $y_{\mathcal{V} \setminus (\mathcal{N}_t \cup \{\ell_t\})}$. Formally:
\begin{equation}
    P(y_t \mid \mathbf{y}_{\mathcal{V} \setminus \{t\}}) = P(y_t \mid \mathbf{y}_{\mathcal{N}_t})
\end{equation}
\end{assumption}

\begin{remark}
This assumption justifies the design of our \textbf{Topology Operator} ($\psi$). It implies that instead of reasoning over the computationally intractable global context, it suffices to construct a high-quality local neighborhood $\mathcal{N}_t$ to theoretically achieve optimal predictive performance.
\end{remark}

\begin{proposition}[Benefit of Spatial Aggregation]
\label{prop:stein}
As established in \cite{stein1956inadmissibility}, in multi-dimensional estimation tasks ($|\mathcal{V}| \ge 3$), independent estimation of node-wise variables is \emph{inadmissible} under quadratic loss. Specifically, there exists a spatially aggregated estimator $\hat{\mathbf{y}}^{\text{agg}}$ (acting as a linear smoother over the neighbor graph) that strictly dominates independent estimation in terms of total Mean Squared Error (MSE):
\begin{equation}
    \mathbb{E}[\|\hat{\mathbf{y}}^{\text{agg}} - \mathbf{y}\|^2]
    <
    \mathbb{E}[\|\hat{\mathbf{y}}^{\text{ind}} - \mathbf{y}\|^2],
\end{equation}
provided the aggregation strength is within a non-trivial bound determined by the graph Laplacian. (proof in Appendix~\ref{proof:mp})
\end{proposition}

\begin{remark}
This proposition formally explains \emph{why} our message passing mechanism is superior to standard independent prompting. By conditioning each prediction on its spatial neighbors, GeoGR\textsuperscript{2} performs structured variance reduction, mitigating the hallucination and bias inherent in isolated LLM predictions.
\end{remark}

\begin{lemma}[Convergence Condition]
\label{lem:convergence}
Let the refinement process be viewed as an operator $T: \mathbb{R}^{|\mathcal{V}|} \to \mathbb{R}^{|\mathcal{V}|}$ acting on a complete metric space. If the Update Operator is designed to act as a contraction mapping with Lipschitz constant $L < 1$, then by the Banach Fixed Point Theorem~\cite{kreyszig1991introductory}, the iterative sequence $\hat{\mathbf{y}}^{(k+1)} = T(\hat{\mathbf{y}}^{(k)})$ converges to a unique fixed point $\mathbf{y}^*$:
\begin{equation}
    \lim_{k \to \infty} \hat{\mathbf{y}}^{(k)} = \mathbf{y}^*.
\end{equation}
\end{lemma}

\begin{remark}
This lemma validates the stability of our \textbf{Update Operator} ($\varphi$). 
\end{remark}

\subsection{Initial State Estimation}
The process initializes with the LLM, $\pi$, which generates a preliminary state estimate $\hat{y}_t^{(0)}$ for each location $\ell_t$. Leveraging the GeoLLM framework~\cite{manvi2024geollm}, $\pi$ utilizes OpenStreetMap data to ground the zero-shot prediction in basic geographic context:
\begin{equation}
    \hat{y}_t^{(0)} = \pi(\texttt{prompt}(\ell_t, \mathcal{T})), \quad \forall \ell_t \in \mathcal{L},
\end{equation}
where $\mathcal{T}$ denotes the task context. While informative, these initial estimates $\hat{y}_t^{(0)}$ are often spatially incoherent, lacking the constraints imposed by neighboring correlations.

\subsection{Dynamic Topology Construction}
To enforce the Spatial Markov Property (Theorem~\ref{asm:markov}), we must explicitly model the interactions between locations. The Topology Operator, $\psi$, acts as a dynamic topology generator. For a target node $\ell_t$, it identifies a neighborhood set $\mathcal{N}_t \subset \mathcal{L} \setminus \{\ell_t\}$ that exerts the strongest spatial influence.

Unlike fixed distance-based kernels in traditional geostatistics, $\psi$ constructs the neighborhood $\mathcal{N}_t$ semantically and spatially:
\begin{equation}
    \mathcal{N}_t = \psi(\ell_t, \mathcal{T}) = \mathcal{N}_t^{\text{prox}} \cup \mathcal{N}_t^{\text{sem}},
\end{equation}
where $\mathcal{N}_t^{\text{prox}}$ comprises the $k$-nearest neighbors ensuring local continuity, and $\mathcal{N}_t^{\text{sem}}$ includes additional reference points selected via LLM reasoning to capture global context or analogous regions. 

\subsection{Semantic Feature Augmentation}
Geographic variables are rarely independent; they are conditioned on latent covariates (e.g., vegetation index influences temperature). The Feature Operator, $\phi$, functions as a feature extractor that maps a location $\ell_t$ to a task-specific subset of auxiliary variables $\mathcal{S}_t \subseteq \mathcal{V}$.
\begin{equation}
    \mathbf{z}_t = \text{Lookup}(\ell_t, \phi(\ell_t, \mathcal{T})),
\end{equation}
where $\mathbf{z}_t \in \mathbb{R}^{|\mathcal{S}_t|}$ represents the values of selected covariates at $\ell_t$. 

\subsection{Natural Language Message Passing}
The core of GeoGR\textsuperscript{2} is the Update Operator, $\varphi$, which mimics the \textit{Aggregate-and-Update} mechanism of GNNs, but operates within the semantic space of an LLM rather than vector space. At iteration $k$, for each target $\ell_t$, the operator aggregates information from its neighborhood $\mathcal{N}_t$. The ``message'' $m_t^{(k)}$ effectively summarizes the spatial context:
\begin{equation}
    m_t^{(k)} = \texttt{Aggregate}\left( \{ (\hat{y}_r^{(k-1)}, \mathbf{z}_r) \mid \ell_r \in \mathcal{N}_t \} \right).
\end{equation}
The Update Operator then performs a non-linear update to the prediction state, conditioned on the current estimate, the aggregated spatial message, and the local features:
\begin{equation}
    \hat{y}_t^{(k)} = \varphi \left( \hat{y}_t^{(k-1)}, m_t^{(k)}, \mathbf{z}_t \right).
\end{equation}
This process allows the LLM to perform ``reasoning by spatial analogy'' adjusting $\hat{y}_t$ to align with the observed values of its neighbors. Driven by the convergence principle (Proposition~\ref{prop:stein}), the refinement continues for $K$ rounds or until spatial consistency is achieved.

\subsection{Algorithm and Iterative Refinement}

The overall procedure of GeoGR\textsuperscript{2} is summarized in Algorithm~\ref{alg:geosr} (Appendix~\ref{app:algorithm}). The process begins with an independent initialization state, $\hat{\mathbf{y}}^{(0)}$, where the base LLM generates zero-shot predictions for each location. 

Instead of relying on traditional numerical solvers, our framework iteratively updates the global prediction state $\hat{\mathbf{y}}^{(k)}$ via natural language reasoning. In each iteration $k$, the Topology Operator $\psi$ and Feature Operator $\phi$ provide the necessary spatial and semantic context for each location. Subsequently, the LLM-based Update Operator $\varphi$ guides each location $\ell_t$ to reconcile its prediction with the gathered neighborhood consensus $\Omega_t$. This refinement mechanism drives a sequential state transition: $\hat{\mathbf{y}}^{(0)} \to \hat{\mathbf{y}}^{(1)} \to \cdots \to \hat{\mathbf{y}}^{(K)}$. By iteratively aligning individual outputs with local context, the system effectively reduces spatial inconsistencies, yielding a final set of predictions $\{\hat{y}_t^{(K)}\}$ that is spatially coherent and grounded in geospatial features.

\section{Experiment}
We evaluate GeoGR\textsuperscript{2} following \citet{manvi2024geollm}, requiring models to generate 10-point scores per task. Performance is measured via Spearman correlation and scoring bias (Eq.~\eqref{eq:bias}), using population density as anchor distribution $\mathbf{d}$ to account for regional imbalances.

We assess four geospatial tasks spanning social and physical geographic data: \textbf{Infant Mortality} (healthcare access, social equity; \cite{manvi2024geollm}, 1950 cities), \textbf{GDP} (economic disparities; \cite{kummu2025downscaled}, 1950 cities), and climate variables \textbf{Temperature} and \textbf{Precipitation} (climate/hydrological patterns; WorldClim\footnote{\url{https://www.worldclim.org/}} \cite{fick2017worldclim}, 2000 global locations each).

\textbf{GeoGR\textsuperscript{2}} builds on LLMs with GeoLLM~\cite{manvi2024geollm} as the \emph{LLM-only baseline} using local context without global topology. We treat GeoLLM outputs as initial predictions $\mathbf{y}_0$ and integrate GeoGR\textsuperscript{2} as a graph-based refinement module to quantify gains from global reasoning. Evaluated LLMs include: \textbf{GPT-3.5-Turbo} \cite{2022gpt} (175B generalist, cutoff Sep 2021), \textbf{GPT-4o-mini} \cite{2024gpt4o} (compact MoE, 10-50B), \textbf{DeepSeek-V3} \cite{2025deepseekv3} (671B MoE), \textbf{GeoGPT} (LLaMA-3.1-70B fine-tuned on 800k geoscience samples), and \textbf{Gemini-3-flash} \cite{2025gemini3flash} (multimodal lightweight, cutoff Dec 2024). Additionally, we include KNN~\cite{bishop2006pattern} and XGBoost~\cite{XGBoost} as traditional baselines under a strict few-shot setting (20 randomly sampled training points) to maintain parity with our training-free approach.
\subsection{Comparison with Generic Agentic Reasoning}
\label{sec:react_baseline}

We implement a ReAct agent~\cite{yao2023react} equipped with geographic retrieval, neighbor lookup, and estimation tools, capped at 20 rounds per location. \cref{tab:react_gpt35} shows severe \textit{convergence instability}: unstructured tool calls exhaust the budget across all tasks, yielding lower Spearman correlation (e.g., 0.599 vs. \textbf{0.747} on Infant Mortality) and higher geographic bias (0.379 vs. \textbf{-0.006}). Token costs reach 11,306--15,739 per location (\textbf{7--10$\times$} GeoGR\textsuperscript{2}). These results highlight a core limitation of generic agents: lacking spatial inductive biases, they incur redundant retrieval without enforcing neighborhood consistency. GeoGR\textsuperscript{2} resolves this by embedding spatial dependencies directly into the inference topology, guaranteeing single-pass execution with lower cost and higher fidelity.

\begin{table}[t]
\centering

\small
\begin{tabular}{@{}lccccc@{}}
\toprule
\textbf{Task} & \textbf{Method} & \textbf{$\rho$} & \textbf{Bias} & \textbf{Tokens/Loc.} \\
\midrule
\multirow{2}{*}{Infant Mort.} & ReAct & 0.599 & 0.379 & 15,739\\
& GeoGR\textsuperscript{2} & \textbf{0.747} & \textbf{-0.006} & \textbf{1,638} \\
\midrule
\multirow{2}{*}{GDP} & ReAct & 0.501 & 0.283  & 12,450 \\
& GeoGR\textsuperscript{2} & \textbf{0.653} & \textbf{0.077} & \textbf{1,744} \\
\midrule
\multirow{2}{*}{Temperature} & ReAct & 0.535 & 0.311 & 11,306 \\
& GeoGR\textsuperscript{2} & \textbf{0.622} & \textbf{0.024} & \textbf{1,591} \\
\midrule
\multirow{2}{*}{Precipitation} & ReAct & 0.457 & 0.140 & 12,579 \\
& GeoGR\textsuperscript{2} & \textbf{0.694} & \textbf{0.039} & \textbf{1,503} \\
\bottomrule
\end{tabular}
\caption{Efficiency and Performance: GeoGR\textsuperscript{2} vs. ReAct Agent (GPT-3.5-Turbo).}
\label{tab:react_gpt35}
\end{table}

\subsection{Main Results}
\begin{table*}[t]
  \centering
  \resizebox{\textwidth}{!}{
  \begin{tabular}{lcccccccc}
  \toprule
  \multirow{2}{*}{\textbf{Model}} & \multicolumn{2}{c}{\textbf{Infant Mortality}} & \multicolumn{2}{c}{\textbf{GDP}} & \multicolumn{2}{c}{\textbf{Temperature}} & \multicolumn{2}{c}{\textbf{Precipitation}} \\
  \cmidrule(lr){2-3} \cmidrule(lr){4-5} \cmidrule(lr){6-7} \cmidrule(lr){8-9}
  & Spearman & Bias & Spearman & Bias & Spearman & Bias & Spearman & Bias \\
  \midrule
  GeoGPT & 0.798 & -0.175 & 0.515 & 0.274 & 0.753 & 0.136 & 0.797 & 0.185 \\
  +GeoGR\textsuperscript{2} & 0.816 (+2.26\%) & -0.031 (-82.3\%) & 0.523 (+1.55\%) & 0.146 (-46.7\%) & 0.771 (+2.39\%) & 0.108 (-20.6\%) & 0.802 (+0.63\%) & 0.157 (-15.1\%) \\
  \midrule
  GPT-3.5-Turbo & 0.445 & -0.188 & 0.512 & 0.618 & 0.462 & 0.074 & 0.594 & 0.031 \\
  +GeoGR\textsuperscript{2} & 0.747 (\underline{+67.87\%}) & \textbf{-0.006} (\underline{-96.8\%}) & \textbf{0.653} (\underline{+27.54\%}) & 0.077 (\underline{-87.5\%}) & 0.622 (+34.63\%) & \textbf{0.024} (\underline{-67.6\%}) & 0.694 (+16.84\%) & 0.039 (+25.8\%) \\
  \midrule
  GPT-4o-mini & 0.704 & -0.142 & 0.565 & 0.295 & 0.310 & 0.203 & 0.564 & 0.103 \\
  +GeoGR\textsuperscript{2} & 0.762 (+8.24\%) & -0.011 (-92.3\%) & 0.585 (+3.54\%) & 0.087 (-70.5\%) & 0.517 (+66.77\%) & 0.085 (-58.1\%) & 0.634 (+12.41\%) & \textbf{0.011} (\underline{-89.3\%}) \\
  \midrule
  DeepSeek-V3 & 0.820 & -0.078 & 0.620 & 0.251 & 0.478 & 0.232 & 0.793 & 0.116 \\
  +GeoGR\textsuperscript{2} & 0.834 (+1.71\%) & -0.026 (-66.7\%) & 0.644 (+3.87\%) & \textbf{0.063} (\underline{-74.9\%}) & 0.563 (+17.78\%) & 0.097 (-58.2\%) & \textbf{0.813} (+2.52\%) & 0.061 (-47.4\%) \\
  \midrule
  Gemini-3-flash & 0.870 & -0.456 & 0.522 & 0.220 & \textbf{0.806} & 0.147 & 0.769 & 0.097 \\
  +GeoGR\textsuperscript{2} & \textbf{0.879} (+1.03\%) & -0.061 (-86.6\%) & 0.535 (+2.49\%) & 0.172 (-21.8\%) & 0.805 (-0.12\%) & 0.143 (-2.72\%) & 0.785 (+2.08\%) & 0.115 (+18.65\%) \\
  \midrule
  KNN     & 0.352 & / & 0.347 & / & 0.431 & / & 0.399 & / \\
  XGBoost & 0.415 & / & 0.460 & / & 0.497 & / & 0.510 & / \\
  \bottomrule
  \end{tabular}}
  \caption{Performance Gains with GeoGR\textsuperscript{2}. Baseline Spearman correlation (higher is better) and Bias (lower absolute value is better) are shown, with percentage increase in Spearman and percentage decrease in Bias for GeoGR\textsuperscript{2}-enhanced models.}
  \label{tab:comparison}
\end{table*}

\cref{tab:comparison} summarizes the performance gains achieved by GeoGR\textsuperscript{2} across various LLM backbones. Our framework consistently improves Spearman correlation while significantly reducing scoring bias (e.g., a \textbf{96.8\%} reduction for GPT-3.5). The following key insights emerge:

\paragraph{Comparison with Traditional Baselines.} Traditional machine learning models, KNN and XGBoost, serve as competitive baselines for capturing local spatial patterns. While vanilla LLMs (e.g., GPT-3.5, GPT-4o-mini) often underperform these numerical models in tasks like GDP and Infant Mortality, the integration of GeoGR\textsuperscript{2} enables even lightweight or general-purpose LLMs to significantly surpass these traditional methods. This highlights the framework's ability to bridge the gap between structured spatial regression and linguistic reasoning.

\paragraph{Scaling and Retrofitting Effects.} GeoGR\textsuperscript{2} demonstrates a robust ``retrofitting'' effect across different model scales. Weaker models like GPT-3.5-Turbo exhibit the most dramatic relative improvements, while state-of-the-art models like DeepSeek-V3 and Gemini-3-flash achieve new performance ceilings. Notably, for GeoGPT---which is already fine-tuned on 800k geoscience samples---GeoGR\textsuperscript{2} further refines its output calibration, suggesting that our retrieval-reasoning mechanism is complementary to domain-specific pretraining.

\paragraph{Mitigating Geographic Bias.} A critical contribution of GeoGR\textsuperscript{2} is the drastic reduction in absolute bias. Even high-performing models like Gemini-3-flash show severe geographic disparities at baseline (e.g., -0.456 on infant mortality). By utilizing population density as an anchor distribution, GeoGR\textsuperscript{2} successfully mitigates data-driven biases, ensuring that spatial reasoning remains equitable across diverse regions regardless of their representation in the original training corpus.

GeoGR\textsuperscript{2} significantly reduces bias in nearly all settings, even achieving near-zero  values in some cases (e.g., \textbf{-0.006} for GPT-3.5-Turbo on infant mortality, \textbf{-0.011} for GPT-4o-mini). For cutting-edge lightweight models like Gemini-3-flash that already possess strong inherent spatial reasoning, GeoGR\textsuperscript{2} still yields notable fairness optimization—reducing its infant mortality bias by 86.6\% with a marginal accuracy boost—with only negligible performance fluctuations on individual tasks that do not compromise overall model performance. While domain-adapted models like GeoGPT begin with strong priors and relatively fair outputs, general-purpose models—particularly those with smaller capacities or outdated training data—stand to gain the most from GeoGR\textsuperscript{2}. This pattern highlights a key strength of our approach: \textbf{GeoGR\textsuperscript{2} offers consistent and interpretable benefits across diverse model types}, enabling both accuracy and equity improvements without requiring changes to the underlying model architecture or retraining procedures.

\begin{figure*}[t] 
    \centering 
    \includegraphics[width=0.9\textwidth]{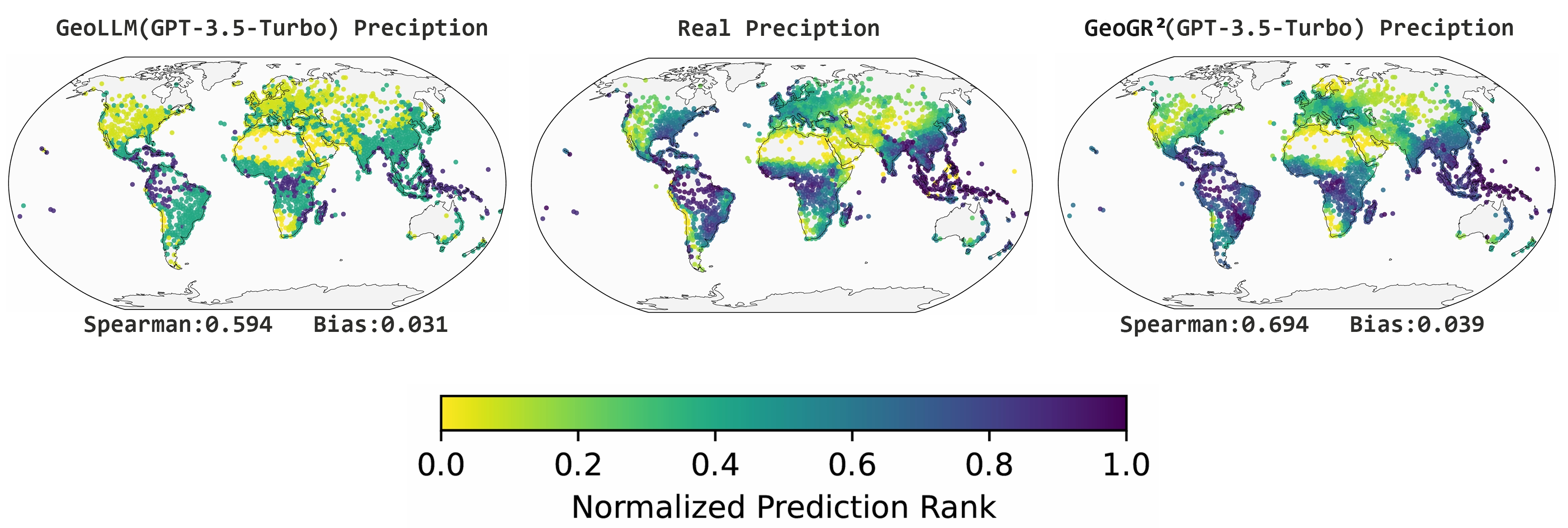} 
    \caption{Spatial prediction rank maps for Precipitation prediction using GPT-3.5-Turbo. \textbf{Middle}: Ground-truth normalized Precipitation ranks. \textbf{Left}: Predictions from the base GeoLLM. \textbf{Right}: Predictions from GeoGR\textsuperscript{2}-enhanced model.} 
    \label{fig5}
    \vspace{-10pt}
\end{figure*}

\cref{fig5} provides a qualitative comparison of spatial precipitation predictions using GPT-3.5-Turbo. From the comparison, we observe that GeoGR\textsuperscript{2} improves the spatial alignment of predicted precipitation distributions. Specifically, the prediction map on the right more closely resembles the ground-truth middle panel, particularly in regions with historically high bias in LLM outputs such as \textbf{Africa}, \textbf{South America} and \textbf{Southeast Asia}. In the baseline GeoLLM output (left), many regions in those regions are consistently over- or under-estimated. These spatial visualizations further support our claim that GeoGR\textsuperscript{2} can effectively correct for structural inequities in LLM outputs, especially for regions that are underrepresented in training data.

\paragraph{Comparison with Strong Spatial Baselines.}
\label{sec:strong_baselines}
To validate that GeoGR\textsuperscript{2}'s gains stem from LLM-based geographical reasoning rather than mere spatial smoothing, we compare it against 7 established spatial methods trained with $N=100$ labeled samples: Kriging, Gaussian Process (GP), Random Forest with coordinates (RF+coord), Kernel Ridge Regression (KRR), Graph Convolutional Network (GCN), Geographically Weighted Regression (GWR), and Graph Smooth. 

\begin{table}[h]
\centering
\caption{GeoGR\textsuperscript{2} (Zero-Shot, $N=0$) vs. Strong Spatial Baselines ($N=100$). Spearman correlation reported.}
\label{tab:strong_baselines}
\small
\begin{tabular}{@{}lcccc@{}}
\toprule
\textbf{Method} & \textbf{Infant Mort.} & \textbf{GDP} & \textbf{Temp.} & \textbf{Precip.} \\
\midrule
Kriging & 0.663 & 0.681 & 0.823 & 0.698 \\
GP (RBF) & 0.752 & 0.695 & 0.879 & 0.749 \\
RF+coord & 0.799 & 0.652 & 0.885 & 0.694 \\
KRR & 0.774 & 0.637 & 0.847 & 0.731 \\
GCN & 0.807 & \textbf{0.707} & \textbf{0.910} & 0.745 \\
GWR & 0.587 & 0.538 & 0.696 & 0.422 \\
GraphSmooth & 0.212 & 0.287 & 0.280 & 0.299 \\
\midrule
\textbf{Gemini-3-flash} & 0.870 & 0.522 & 0.806 & 0.769 \\
\textbf{+ GeoGR\textsuperscript{2}} & \textbf{0.879} & 0.535 & 0.805 & \textbf{0.785} \\
\bottomrule
\end{tabular}
\end{table}

As shown in \cref{tab:strong_baselines}, our zero-shot approach achieves comparable or superior performance to supervised baselines using 100 labeled points. Notably, on the Infant Mortality task (which exhibits weaker spatial agglomeration), GeoGR\textsuperscript{2} significantly outperforms all supervised spatial interpolators (0.879 vs. GCN's 0.807). This confirms that the performance gain is driven by the LLM's semantic reasoning over geographical contexts (e.g., POI semantics, climate covariates), which pure numerical spatial smoothers cannot capture.

\subsection{Ablation Study}

\paragraph{Topology Construction Strategy.} We further investigate the effectiveness of our topology construction strategy, which comprises two components: (i) automatically selecting the ten nearest neighboring points ({GeoGR\textsuperscript{2} w/o${N}_t^{\text{sem}}$}), and (ii) selecting supplementary points dynamically by the LLM ({GeoGR\textsuperscript{2} w/o  ${N}_t^{\text{prox}}$}). We evaluate these variants using GPT-3.5-Turbo as the base model, on the same datasets as in the main experiments. As reported in \cref{tab:point}, the full {GeoGR\textsuperscript{2}} model achieves the highest Spearman correlation on all tasks except for temperature, where the {GeoGR\textsuperscript{2} w/o ${N}_t^{\text{sem}}$} variant slightly outperforms it (0.668 vs. 0.644). This suggests that supplementary points chosen by LLM contribute marginally to performance in this specific case, likely due to the strong spatial autocorrelation of temperature data. Importantly, the performance drop is minor, demonstrating the robustness of the full model. In terms of fairness, removing the nearest-neighbor selection ({GeoGR\textsuperscript{2} w/o ${N}_t^{\text{prox}}$}) only slightly reduces bias. For example, in the infant mortality task, the absolute bias decreases by just {0.02}, indicating that the inclusion of the nearest points plays a limited role in mitigating bias.
We further observe that GeoGR\textsuperscript{2} w/o ${N}_t^{\text{sem}}$ consistently achieves higher Spearman correlations than {GeoGR\textsuperscript{2} w/o ${N}_t^{\text{prox}}$} across tasks. This result reinforces \textit{Tobler’s First Law of Geography}, which posits that "everything is related to everything else, but near things are more related than distant things."

\begin{table}[h]
\centering

\scriptsize  
\begin{tabular}{@{}llcccc@{}}
\toprule
\multicolumn{2}{c}{Method Variant} & \multicolumn{4}{c}{Tasks} \\
\cmidrule(lr){1-2} \cmidrule(lr){3-6}
Method & Metric & \shortstack{Infant mort.} & GDP & \shortstack{Temp.} & \shortstack{Precip.} \\
\midrule
\multirow{2}{*}{\shortstack[l]{GeoGR\textsuperscript{2} w/o\\ ${N}_t^{\text{prox}}$}} 
    & Spearman &0.645 &0.577 &0.640 &0.659 \\
    & Bias     &0.034 &0.198 &0.038 &0.051 \\
\addlinespace[0.2em]
\multirow{2}{*}{\shortstack[l]{GeoGR\textsuperscript{2} w/o\\ ${N}_t^{\text{sem}}$}} 
    & Spearman &0.727 &0.650 &\textbf{0.668} &0.653 \\
    & Bias     &0.004 &0.064 &0.031 &0.027 \\
\addlinespace[0.2em]
\multirow{2}{*}{GeoGR\textsuperscript{2}} 
    & Spearman &\textbf{0.747} &\textbf{0.653} &0.644 &\textbf{0.694} \\
    & Bias    &-0.006 &0.077 &0.024 &0.039 \\
\bottomrule
\end{tabular}
\caption{Ablation Study on Point Selection Strategy}
\label{tab:point}
\end{table}

\paragraph{Feature Operator.}

To evaluate the impact of covariates selected by the {Feature Operator}, we conduct an ablation study comparing the full {GeoGR\textsuperscript{2}} model with a variant that excludes these external covariates. As shown in \cref{tab:bio}, incorporating bioclimatic variables consistently improves performance across most tasks. For instance, in the infant mortality task, the Spearman correlation increases from 0.705 to 0.747, alongside a reduction in absolute bias. Similar gains are observed for GDP and precipitation tasks. Interestingly, for the temperature task, excluding bioclimatic covariates results in a marginal Spearman increase, but the improvement is negligible and likely reflects feature redundancy rather than added value—since the target itself is temperature, additional temperature-related inputs may contribute limited new information. 

\begin{table}[h]
\centering

\scriptsize  
\begin{tabular}{@{}llcccc@{}}
\toprule
\multicolumn{2}{c}{Method Variant} & \multicolumn{4}{c}{Tasks} \\
\cmidrule(lr){1-2} \cmidrule(lr){3-6}
Method & Metric & \shortstack{Infant mort.} & GDP & \shortstack{Temp.} & \shortstack{Precip.} \\
\midrule
\multirow{2}{*}{\shortstack[l]{GeoGR\textsuperscript{2} w/o\\ external variables}} 
  & Spearman &0.705 &0.650 &\textbf{0.649} &0.691 \\
    & Bias     &-0.019 &0.027 &0.022 &0.041 \\
\addlinespace[0.2em]
\multirow{2}{*}{GeoGR\textsuperscript{2}} 
    & Spearman &\textbf{0.747} &\textbf{0.653} &0.644 &\textbf{0.694} \\
    & Bias    &-0.006 &0.077 &0.024 &0.039 \\
\bottomrule
\end{tabular}
\caption{Ablation Study on External Variables}
\label{tab:bio}
\end{table}

\paragraph{Effect of Iterative Refinement Rounds.} We evaluate how refinement rounds impact GeoGR\textsuperscript{2} using DeepSeek-V3 across four tasks. Spearman correlation improves initially, typically peaking at rounds 2--3. Beyond this, additional iterations yield marginal gains or slight degradation, likely due to capacity saturation or overfitting to local structures. Conversely, geographic bias steadily decreases with further refinement, revealing an accuracy--fairness trade-off: later rounds prioritize equity over fidelity by reducing regional output disparities. For GDP and infant mortality, Spearman slightly declines post-peak while bias drops monotonically. We hypothesize that extended refinement induces a conservative prediction regime, narrowing the output distribution and mitigating regional variance. This aligns with fairness-oriented optimization, trading minor precision losses for more equitable geographic treatment.

\begin{figure}[t] 
    \centering 
    \includegraphics[width=0.49\textwidth]{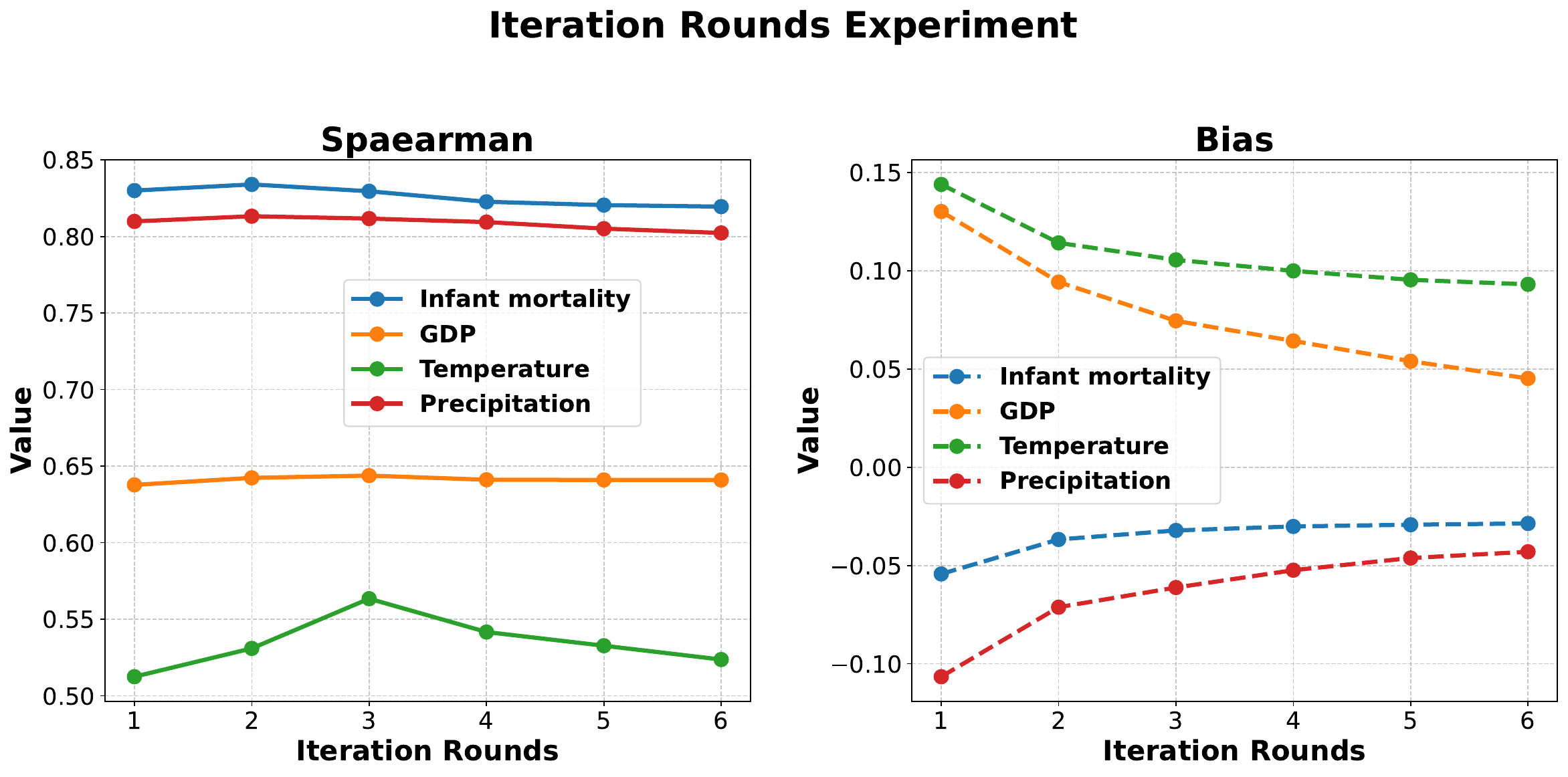} 
    \caption{Effect of iterative refinement rounds on model performance.}
    \label{fig4}
\end{figure}

\begin{figure}[t]
    \centering
    \includegraphics[width=0.45\textwidth]{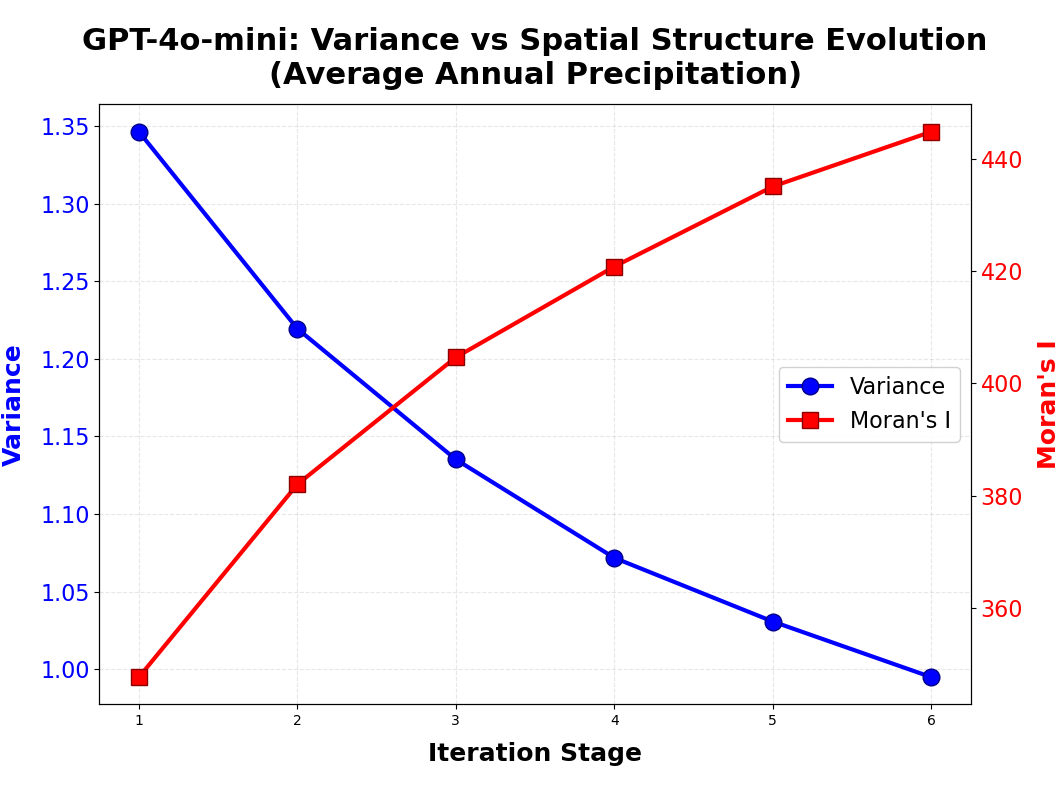}
    \caption{Spatial heterogeneity and convergence metrics across refinement rounds (Precipitation task, GPT-4o-mini). }
    \label{fig:moran}
\end{figure}

\paragraph{Spatial Structure Preservation.}
Iterative refinement reduces prediction variance while \textbf{increasing} Moran's $I$ (\cref{fig:moran}), contradicting over-smoothing where spatial heterogeneity collapses. This indicates that GeoGR\textsuperscript{2} amplifies true spatial signals: estimates converge toward \emph{spatially analogous} neighbors per Tobler's First Law, not a homogeneous average. The precipitation task exemplifies this---coastal vs. continental rainfall gradients sharpen as initial noise suppresses.

\section{Application: Disaster Prediction}
\begin{figure}[t] 
 \centering 
    \includegraphics[width=0.5\textwidth]{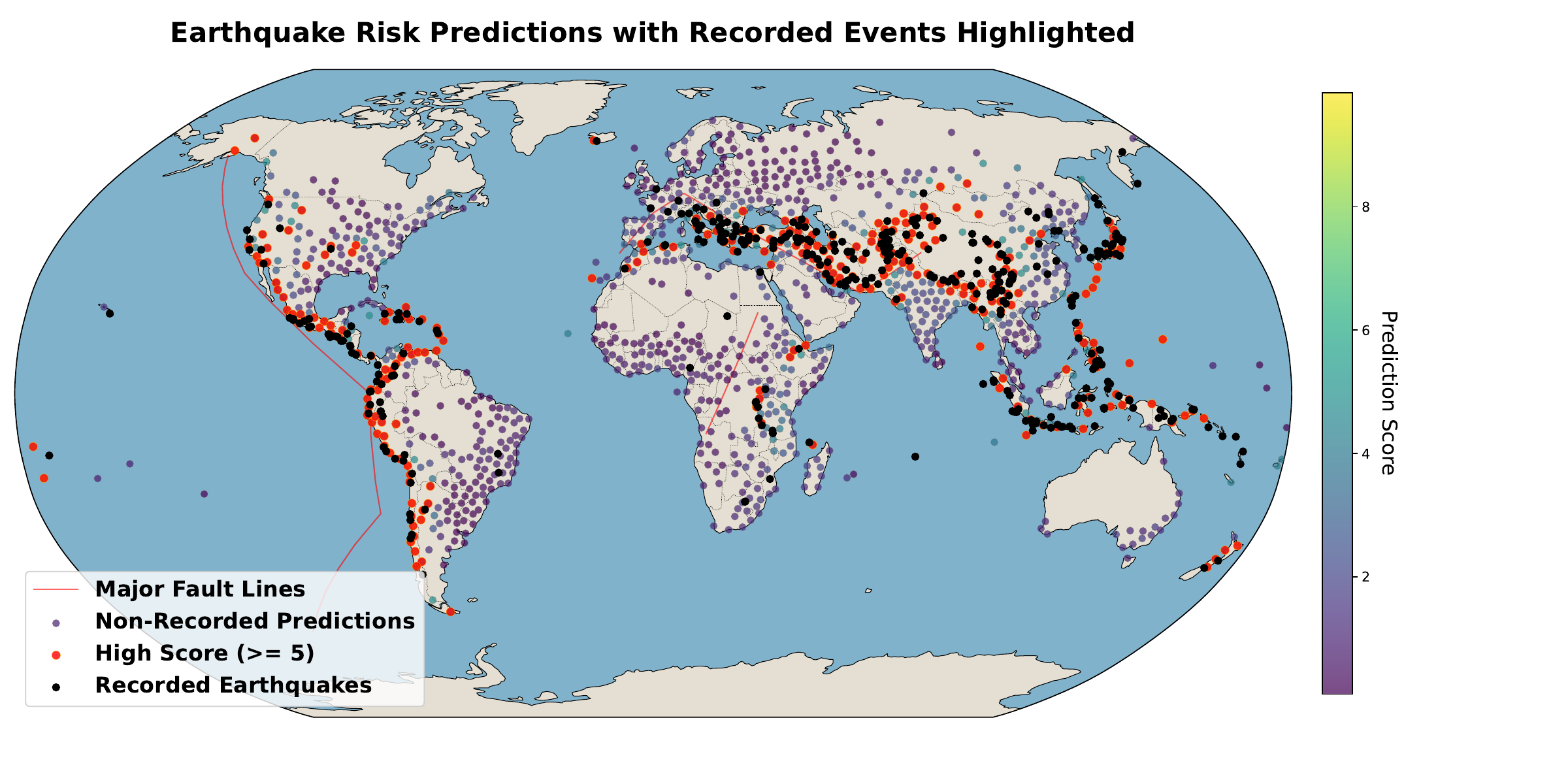} 
    \caption{High Earthquake Risk Given by GeoGR²} 
    \label{fig7}
\end{figure}

\textbf{GeoGR\textsuperscript{2}} applies to diverse scenarios including urban planning, disaster prediction, and community surveys. We compiled global seismic records (2000–2025) covering \textbf{374 cities with earthquake events}, supplemented by 1,091 randomly selected cities, yielding 1,465 cities evaluated via \textbf{GeoGR\textsuperscript{2}} (Gemini‑3‑flash) for seismic risk. \cref{fig7} shows the spatial distribution: red-shaded areas denote tectonic plate boundaries (historically seismically active), black markers indicate cities with recorded earthquakes, and red markers highlight model-identified high-risk cities. As shown, \textbf{GeoGR\textsuperscript{2}} accurately identifies cities facing elevated seismic risk.

\section{Conclusion}

We presented GeoGR\textsuperscript{2}, a training-free framework that operationalizes LLMs' latent capacity for spatial correlation understanding through geostatistically-motivated iterative refinement. By dynamically constructing spatial neighborhoods and performing natural language message passing, GeoGR\textsuperscript{2} enforces spatial consistency without altering model parameters. Extensive experiments across physical and socioeconomic tasks demonstrate state-of-the-art zero-shot accuracy alongside drastic reductions in geographic bias. Future work will extend this paradigm to spatiotemporal forecasting and multi-scale spatial hierarchies.

Beyond geospatial applications, this work offers broader insights for the LLM community: it demonstrates how non-linguistic structural priors can be effectively injected into LLM inference via natural-language message passing and structured prompting, providing a new paradigm for structured reasoning without parameter updates.


\section*{Limitations}
\textbf{Iteration Stopping without Ground Truth.} 
Operating in a zero-shot regime, our framework lacks access to 
ground-truth labels during inference, precluding objective 
selection of the optimal iteration count $K$. The $K$ values 
reported in our experiments (e.g., $K=2$ or $K=3$) are 
determined through post-hoc analysis on the full dataset, 
which is unavailable at deployment time. Future work should 
explore unsupervised stopping criteria, such as monitoring 
spatial consistency convergence or stabilization of 
prediction deltas across rounds, to determine termination 
without labeled data.

\textbf{Temporal Dimension Extension.} 
The present work is confined to spatial prediction tasks. Incorporating temporal dynamics would significantly broaden applicability, enabling spatiotemporal forecasting for critical domains such as climate trend projection, urban growth modeling, and epidemic spread prediction. Subsequent research should focus on extending the operator design to accommodate time-evolving neighborhood structures and sequential covariate dependencies.

\section*{Acknowledgments}
This work was supported in part by the National Natural Science Foundation of China under Grant No. 62406206, and by the Fundamental Research Funds for the Central Universities.
\nocite{langley00}
\bibliography{custom}

\newpage
\appendix

\section{Proof of Proposition~\ref{prop:stein}}
\label{proof:mp}
\begin{proof}
Let the ground truth be $\mathbf{y} \in \mathbb{R}^n$ ($n = |\mathcal{V}|$) and the independent zero-shot predictions be $\hat{\mathbf{y}}^{\text{ind}} = \mathbf{y} + \boldsymbol{\epsilon}$, where $\boldsymbol{\epsilon} \sim \mathcal{N}(\mathbf{0}, \sigma^2 \mathbf{I})$. The risk (MSE) of the independent estimator is $R(\hat{\mathbf{y}}^{\text{ind}}) = n\sigma^2$.

The GeoGR\textsuperscript{2} refinement acts as a linear smoother $\mathbf{S} = (\mathbf{I} - \lambda \mathbf{L})$, where $\mathbf{L}$ is the symmetric normalized Laplacian and $\lambda > 0$ is the step size. The new estimator is $\hat{\mathbf{y}}^{\text{agg}} = \mathbf{S}\hat{\mathbf{y}}^{\text{ind}}$. The exact risk of this estimator is the sum of squared bias and variance:
\begin{equation}
    R(\hat{\mathbf{y}}^{\text{agg}}) = \|\underbrace{(\mathbf{S} - \mathbf{I})\mathbf{y}}_{\text{Bias}}\|^2 + \text{tr}(\underbrace{\mathbf{S} \text{Cov}(\hat{\mathbf{y}}^{\text{ind}}) \mathbf{S}^T}_{\text{Variance}}).
\end{equation}

Substituting $\mathbf{S} - \mathbf{I} = -\lambda \mathbf{L}$:
\begin{align}
    \Delta R &= R(\hat{\mathbf{y}}^{\text{agg}} - R(\hat{\mathbf{y}}^{\text{ind}}) \\
    &= \|-\lambda \mathbf{L}\mathbf{y}\|^2 + \sigma^2 \text{tr}((\mathbf{I} - \lambda \mathbf{L})^2) - n\sigma^2 \\
    &= \lambda^2 \|\mathbf{L}\mathbf{y}\|^2 + \sigma^2 \text{tr}(\mathbf{I} - 2\lambda \mathbf{L} + \lambda^2 \mathbf{L}^2) - n\sigma^2 \\
    &= \lambda^2 \|\mathbf{L}\mathbf{y}\|^2 - 2\lambda \sigma^2 \text{tr}(\mathbf{L}) + \lambda^2 \sigma^2 \text{tr}(\mathbf{L}^2).
\end{align}

We group terms by powers of $\lambda$:
\begin{equation}
    \Delta R = \underbrace{\lambda^2 \left( \|\mathbf{L}\mathbf{y}\|^2 + \sigma^2 \text{tr}(\mathbf{L}^2) \right)}_{\text{Penalty (Quadratic, Positive)}} - \underbrace{2\lambda \sigma^2 \text{tr}(\mathbf{L})}_{\text{Gain (Linear, Positive)}}.
\end{equation}

Since $\text{tr}(\mathbf{L}) > 0$ for any valid graph structure, the linear reduction term dominates the quadratic penalty term for small $\lambda$. Specifically, $\Delta R < 0$ holds strictly whenever:
\begin{equation}
    0 < \lambda < \frac{2\sigma^2 \text{tr}(\mathbf{L})}{\|\mathbf{L}\mathbf{y}\|^2 + \sigma^2 \text{tr}(\mathbf{L}^2)}.
\end{equation}
Thus, there always exists a non-trivial range of $\lambda$ where GeoGR\textsuperscript{2} strictly dominates the independent estimator.
\end{proof}

\section{On the Distinction from General LLM Graph Reasoning}
\addcontentsline{toc}{section}{On the Distinction from General LLM Graph Reasoning}

We clarify that GeoGR\textsuperscript{2} is fundamentally a \textbf{geospatial prediction framework}, not a contribution to general-purpose LLM graph reasoning. Standard LLM graph reasoning benchmarks (e.g., node classification on citation networks, link prediction on knowledge graphs) assume a \textit{given, static graph topology} and evaluate the LLM's ability to reason over explicit relational structures~\cite{zhang2023graph,fatemi2023talk,wang2023can,chen2023label,shang2025survey}. In contrast, GeoGR\textsuperscript{2} addresses a \textbf{zero-shot regression task over latent spatial graphs}, where (1) the graph topology is unknown and must be dynamically discovered from geographic coordinates, (2) node features are natural language descriptions of locations, and (3) the objective is to estimate continuous geospatial variables rather than classify discrete graph properties. Due to this mismatch in task formulation, input modality, and evaluation protocol, direct comparison with general LLM graph reasoning methods is neither feasible nor informative. The graph formalism in GeoGR\textsuperscript{2} serves exclusively as a mathematical abstraction for spatial autocorrelation, consistent with classical geostatistical practice~\cite{cressie1993statistics,besag1974spatial}.

\section{Algorithm Details}
\label{app:algorithm}

\begin{algorithm}[H]
\small
\setlength{\algorithmicindent}{1.2em}
\caption{GeoGR\textsuperscript{2}: Iterative Spatial Refinement via Dynamic Neighborhood Aggregation}
\label{alg:geosr}
\begin{algorithmic}[1]
\REQUIRE Locations $\mathcal{L}$, Task $\mathcal{T}$, Iterations $K$
\ENSURE Refined Predictions $\{\hat{y}_t^{(K)}\}$
\STATE \textbf{Initialization:} $\hat{y}_t^{(0)} \gets \pi(\ell_t, \mathcal{T})$ for all $\ell_t \in \mathcal{L}$
\FOR{$k = 1$ to $K$}
    \STATE \textcolor{gray}{// Parallel Update Step}
    \FORALL{$\ell_t \in \mathcal{L}$}
        \STATE \textbf{Topology:} Construct neighborhood $\mathcal{N}_t \gets \psi(\ell_t, \mathcal{T})$
        \STATE \textbf{Features:} Retrieve covariates $\mathbf{z}_t, \{\mathbf{z}_r\}_{r \in \mathcal{N}_t}$ via $\phi$
        \STATE \textbf{Message Passing:} 
        \STATE \quad Gather neighbor states $\Omega_t = \{\hat{y}_r^{(k-1)} \mid \ell_r \in \mathcal{N}_t\}$
        \STATE \quad Update $\hat{y}_t^{(k)} \gets \varphi(\hat{y}_t^{(k-1)}, \Omega_t, \mathbf{z}_t, \{\mathbf{z}_r\})$
    \ENDFOR
\ENDFOR
\STATE \textbf{return} $\{\hat{y}_t^{(K)} \mid \ell_t \in \mathcal{L}\}$
\end{algorithmic}
\end{algorithm}
\vspace{-0.5em}
\noindent\footnotesize{\textit{Note:} The neighborhood construction function $\psi(\ell_t, \mathcal{T})$ selects 15 reference points through a hybrid strategy: (1) 10 nearest neighbors computed using Haversine distance; (2) up to 5 additional points selected by an LLM from candidates within a 5°×5° geographic window based on prediction diversity and spatial pattern revelatory power.}

\section{Application}
In this study, we evaluate the earthquake risk prediction capability of \textbf{GeoGR²} against a baseline \textbf{GeoLLM} (both implemented on Gemini-3-flash). Using global seismic records from 2000 to 2025, we identified \textbf{374 cities} with documented earthquake events. To examine model robustness across varying data scales, we constructed three datasets: a \textbf{1:1 balanced set} (374 recorded cities + 374 randomly selected cities without records), a \textbf{1:2 extended set} (374 recorded + 748 non-recorded), and a \textbf{full set of 1,465 cities} (374 recorded + 1,091 randomly sampled globally). Each city was assigned an earthquake risk score (0–10) via an initial \textbf{GeoLLM} prediction based on its name, coordinates, and task description. \textbf{GeoGR²} then refined these scores through topology construction, covariate feature selection, and iterative value updating. A city was classified as high-risk if its score ranked in the top 50\% of all evaluated cities; prediction was considered a true positive if the city had a recorded earthquake.

As detailed in Tables \ref{tab:metrics_comparison} and \ref{tab:performance_comparison}, \textbf{GeoGR²} consistently outperforms \textbf{GeoLLM} across all configurations. Notably, on the full dataset, \textbf{GeoGR²} improves precision from 0.4198 to \textbf{0.4531}, recall from 0.8743 to \textbf{0.9171}, accuracy from 0.6594 to \textbf{0.6962}, F1-score from 0.5672 to \textbf{0.6065}, and AUC-ROC from 0.8160 to \textbf{0.8480}. These gains highlight \textbf{GeoGR²}'s enhanced capacity to integrate structural and covariate information, leading to more reliable and calibrated risk assessments—especially as data scale increases.

\begin{table}
\centering
\caption{Earthquake Prediction Given by GeoLLM(Gemini-3-flash)}
\label{tab:metrics_comparison}
\scriptsize

\begin{tabular}{@{}lcccc@{}}
\toprule
\multirow{2}{*}{Metrics} & \multicolumn{3}{c}{Data Configuration} \\
\cmidrule(l){2-4}
 & 1:1 Ratio & 1:2 Ratio & Full Dataset \\
 & (748 points) & (1122 points) & (1465 points) \\
\midrule
Precision & 0.7469 & 0.5575 & 0.4198 \\
Recall & 0.8128 & 0.8556 & 0.8743 \\
Accuracy & 0.7687 & 0.7255 & 0.6594 \\
F1-score & 0.7785 & 0.6751 & 0.5672 \\
AUC-ROC & 0.8217 & 0.8184 & 0.8160 \\
\bottomrule
\end{tabular}
\end{table}

\begin{table}
\centering
\caption{Earthquake Prediction Given by GeoGR²(Gemini-3-flash)}
\label{tab:performance_comparison}
\scriptsize
\begin{tabular}{@{}lcccc@{}}
\toprule
\multirow{2}{*}{Metrics} & \multicolumn{3}{c}{Data Configuration} \\
\cmidrule(l){2-4}
 & 1:1 Ratio & 1:2 Ratio & Full Dataset \\
 & (748 points) & (1122 points) & (1465 points) \\
\midrule
Precision & 0.7781 & 0.6011 & 0.4531 \\
Recall & 0.7696 & 0.8895 & 0.9171 \\
Accuracy & 0.7726 & 0.7626 & 0.6962 \\
F1-score & 0.7738 & 0.7174 & 0.6065 \\
AUC-ROC & 0.8335 & 0.8439 & 0.8480 \\
\bottomrule
\end{tabular}
\end{table}

\section{Prompts}

\begin{figure}[htb!] 
    \centering 
    \includegraphics[width=0.49\textwidth]{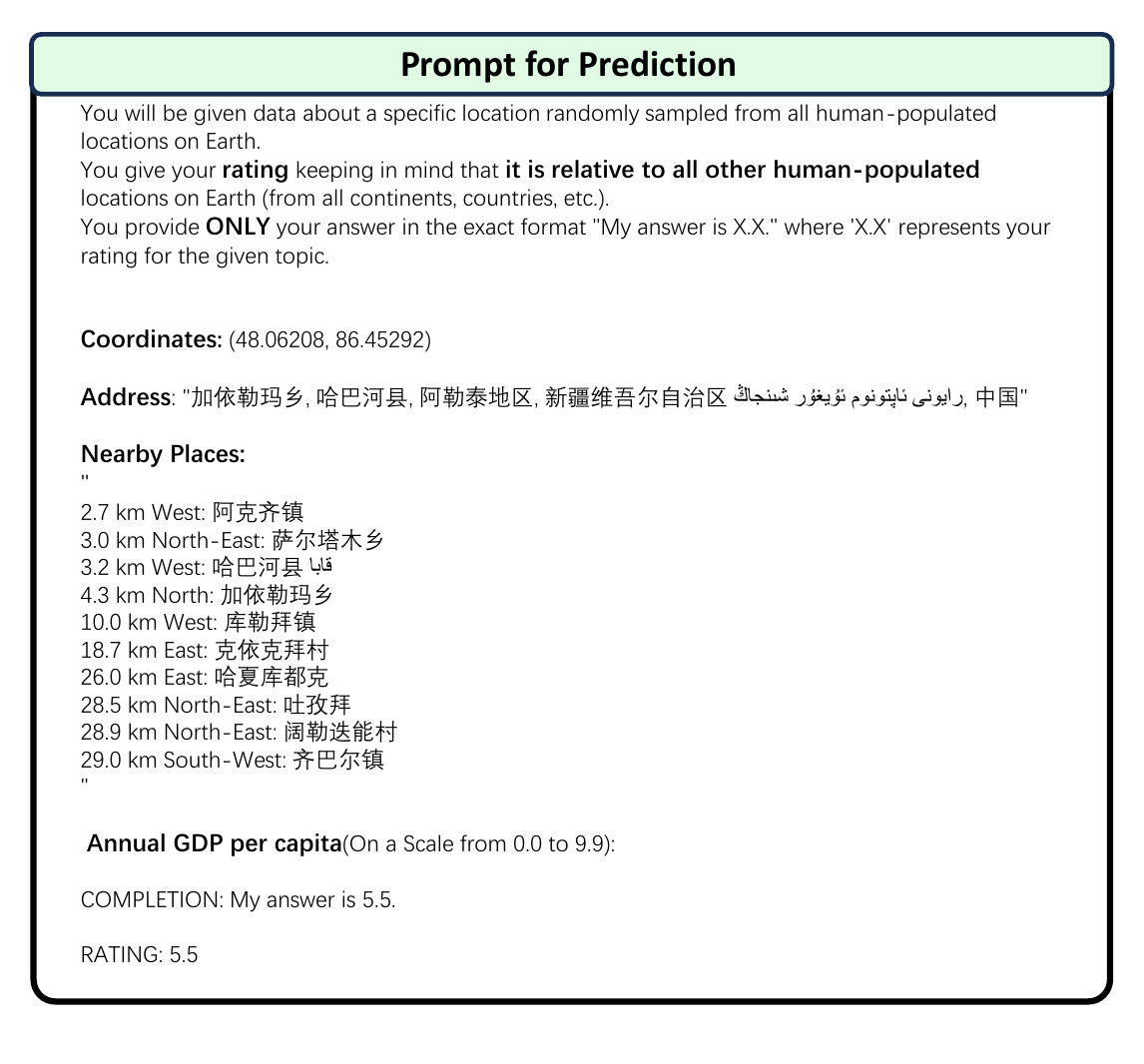} 
    \caption{The prompt for the initial prediction.} 
    \label{app:prediction}
\end{figure}

\begin{figure}[htb!] 
    \centering 
    \includegraphics[width=0.49\textwidth]{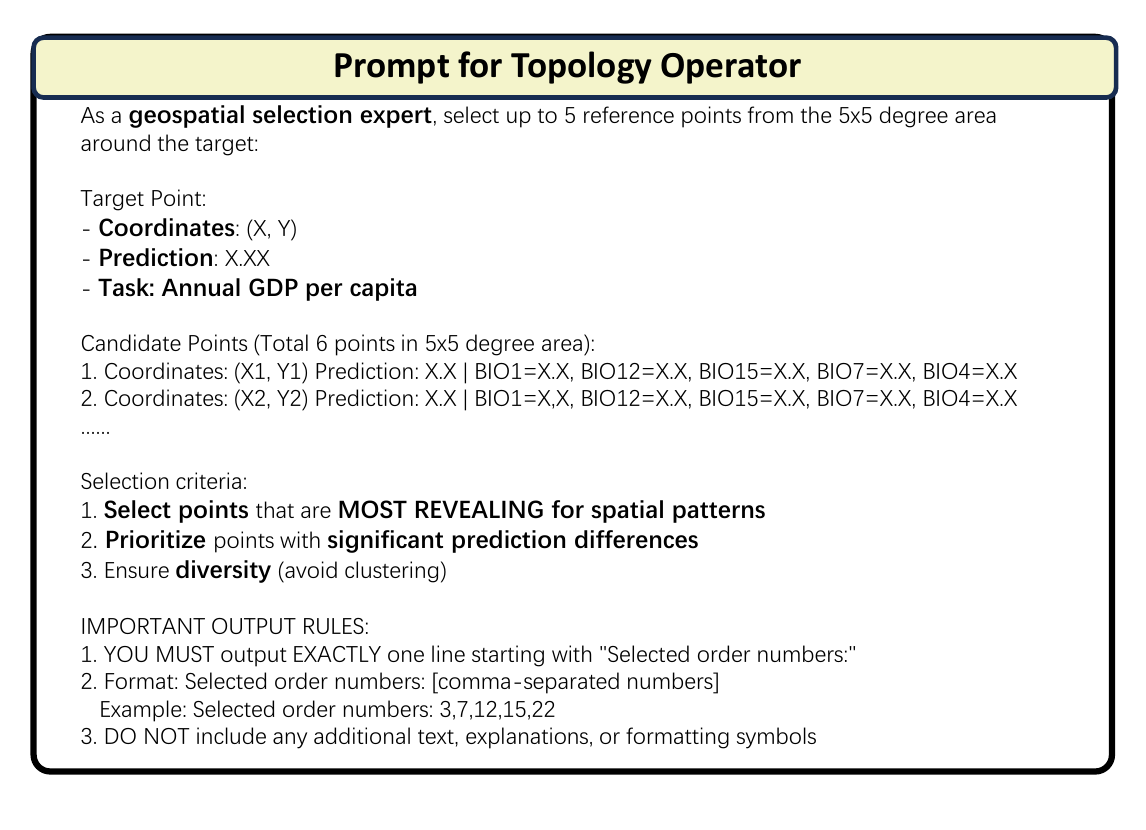} 
    \caption{The prompt for the topology construction.} 
    \label{app:psa}
\end{figure}

\cref{app:prediction}, \cref{app:psa}, \cref{app:vsa} and \cref{app:ra} visually represent the prompt design for the initial, topology construction, covariate selection, and update prediction, illustrating how  topology structure and covariate message are aggregated to optimize geospatial prediction within the GeoGR² framework.

\begin{figure}[htb!] 
    \centering 
    \includegraphics[width=0.49\textwidth]{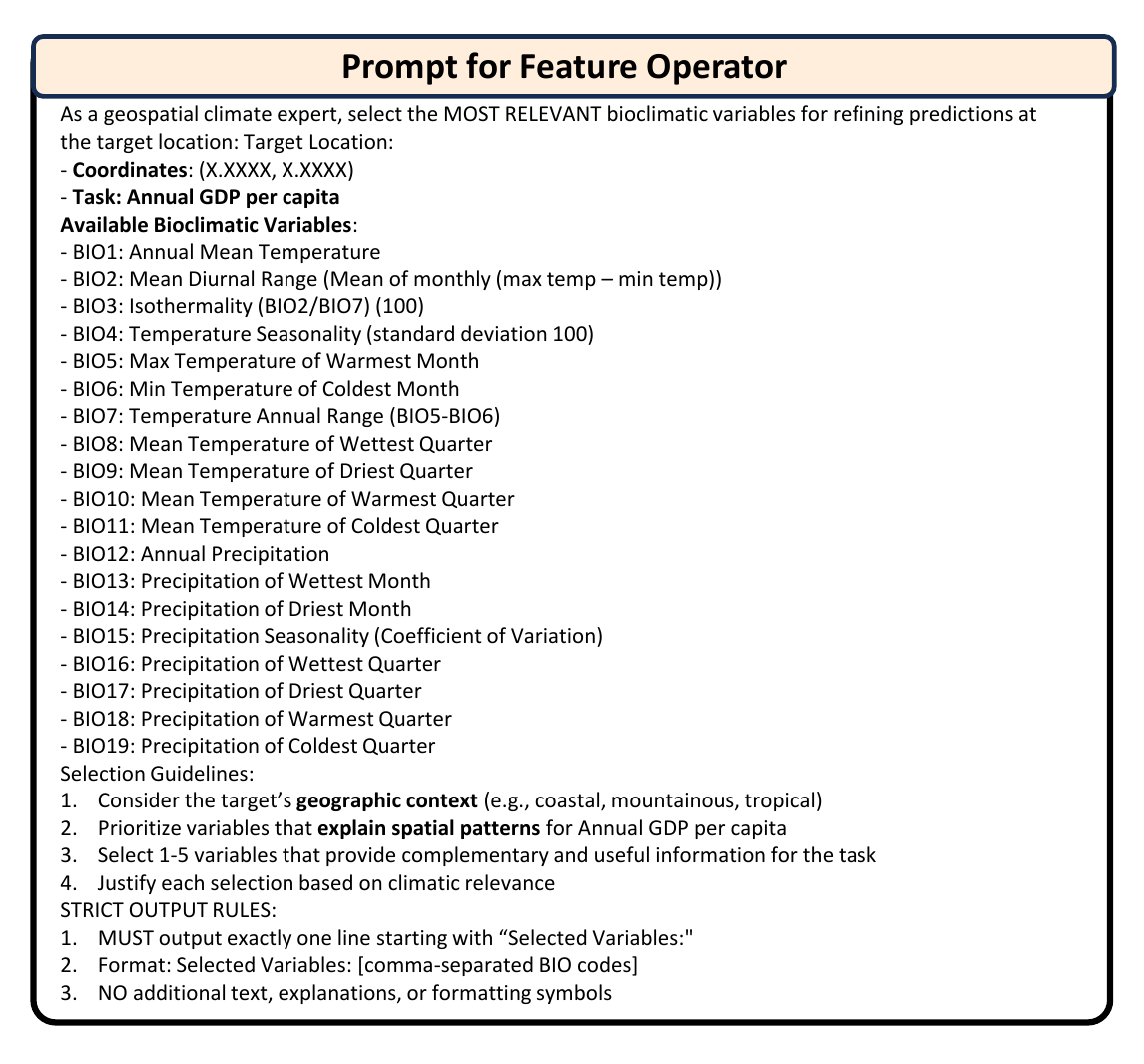} 
    \caption{The prompt for the covariates selection.} 
    \label{app:vsa}
\end{figure}

\begin{figure}[htb!] 
    \centering 
    \includegraphics[width=0.49\textwidth]{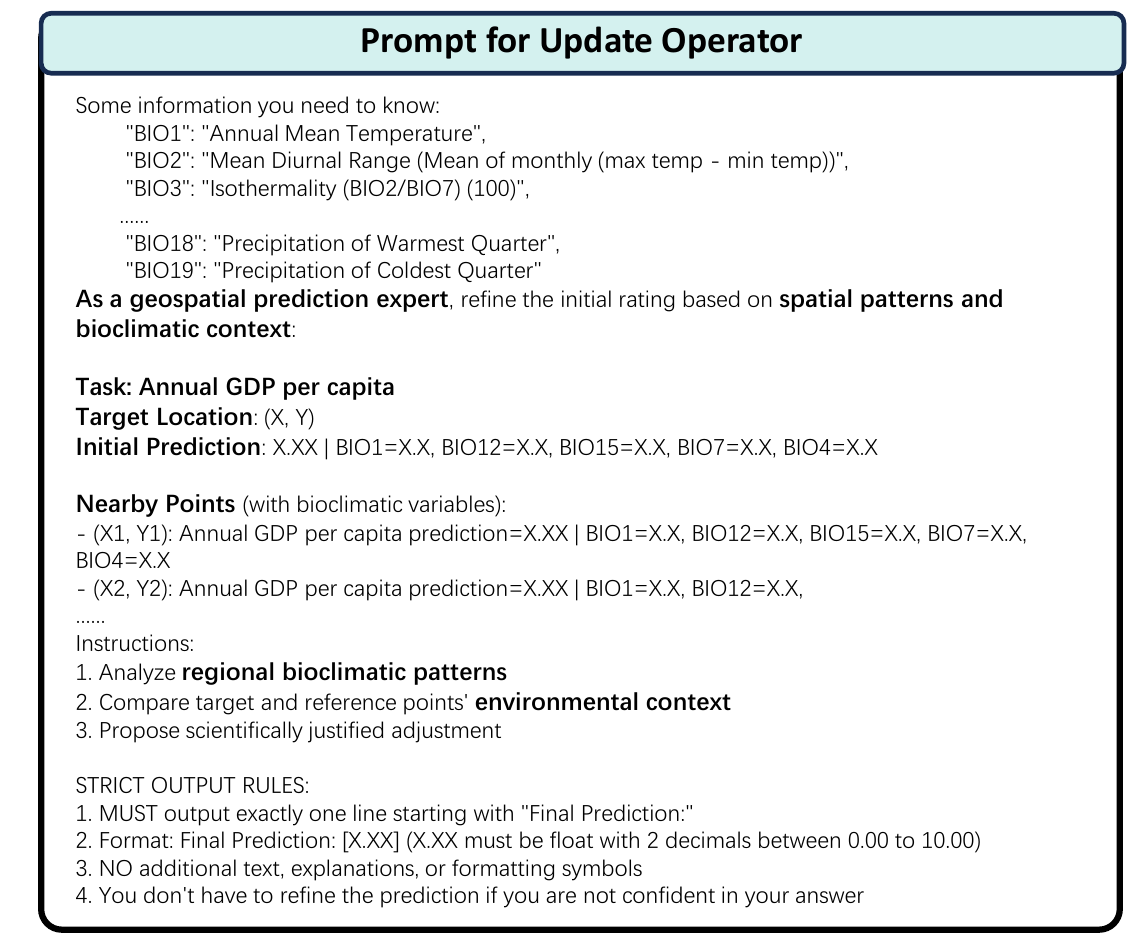} 
    \caption{The prompt for the update prediction.} 
    \label{app:ra}
\end{figure}

\begin{figure}[htb!] 
    \centering 
    \includegraphics[width=0.49\textwidth]{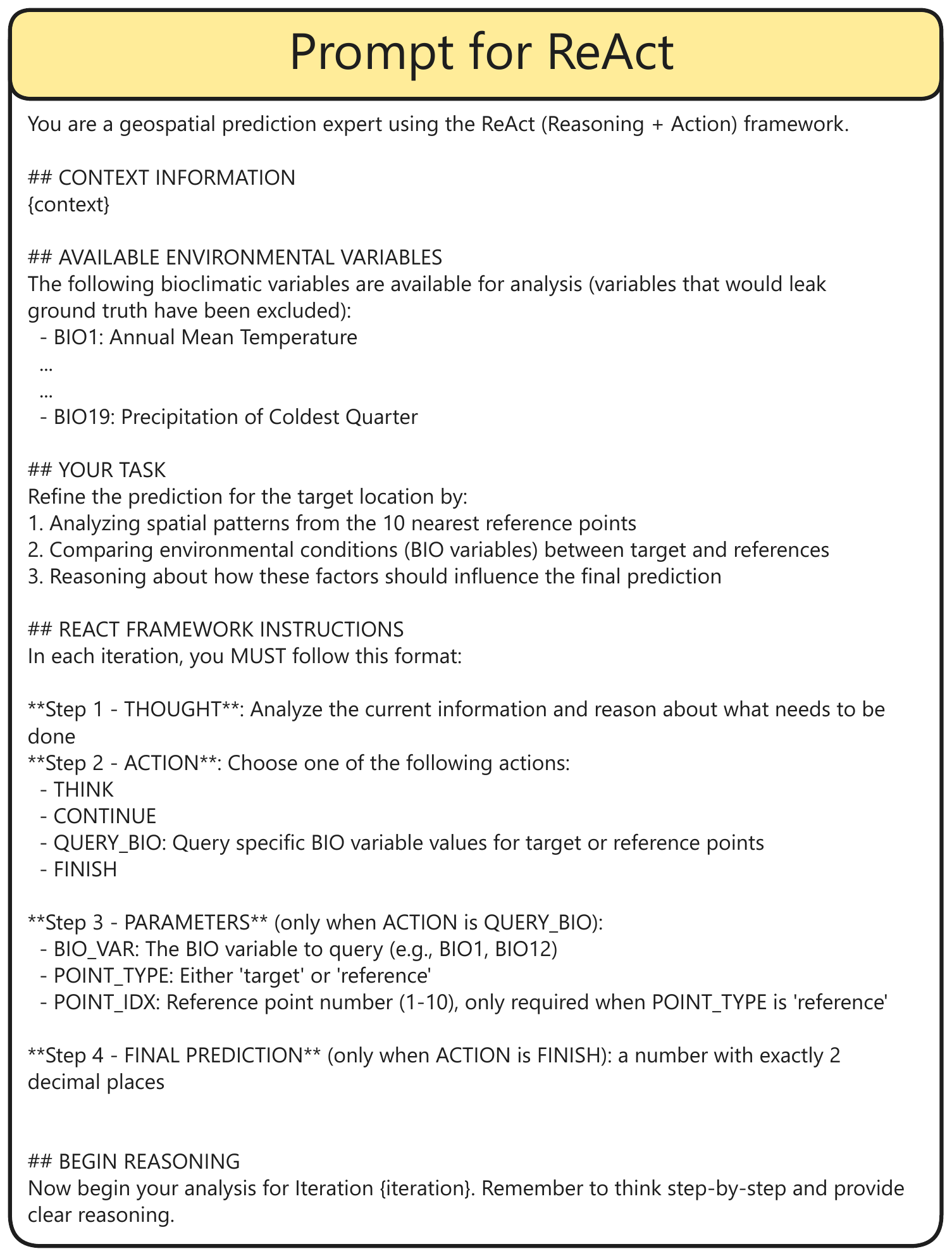} 
    \caption{The prompt for ReAct, one of the baseline methods.}
    \label{app:vsa}
\end{figure}

\section{Token Count}
We measured the token consumption per data point per round, and the results are summarized in \cref{tab:Ogpt_token}.
\begin{table}
\centering
\caption{Token consumption per point per round}
\label{tab:Ogpt_token}
\scriptsize  

\begin{tabular}{@{}lccc@{}}
\toprule
Method & Average &  Min &  Max \\
\midrule
GeoLLM & 311 & 184  &  489\\
GeoGR² & 1,868 &  1,215 & 2,597\\
\bottomrule
\end{tabular}
\end{table}

\section{Use of AI assistant}
We acknowledge the use of large language models (GPT-5.2 and Claude) for language polishing, grammar correction, and stylistic refinement during the preparation of this manuscript. All scientific content, research findings, data analysis, and conclusions were conceived, verified, and approved solely by the authors. The AI tools were employed exclusively for enhancing readability and linguistic clarity, without altering the substantive scientific meaning or intellectual contributions of the work.

\section{Supplementary Experiment}
\subsection{Absolute Numerical Prediction.} 
While our main experiments use 1-10 relative scoring to mitigate hallucination in weaker LLMs, stronger models can directly output calibrated physical units. As shown in \cref{tab:absolute_pred}, Gemini-3-flash + GeoGR\textsuperscript{2} achieves an $R^2$ of 0.787 and MAE of 6.11 on Infant Mortality, vastly outperforming supervised baselines ($N=100$), demonstrating the framework's practical utility for real-world deployment.

\begin{table}[h]
\centering

\small
\begin{tabular}{@{}lcccc@{}}
\toprule
\textbf{Method} & \textbf{Spearman} & \textbf{RMSE} & \textbf{MAE} & \textbf{$R^2$} \\
\midrule
Kriging & 0.663 & 17.00 & 12.39 & 0.438 \\
GP (RBF) & 0.752 & 14.14 & 9.14 & 0.633 \\
GCN & 0.807 & 16.15 & 10.03 & 0.522 \\
\midrule
Gemini-3-Flash & 0.894 & 11.72 & 6.93 & 0.749 \\
\textbf{+ GeoGR\textsuperscript{2}} & \textbf{0.926} & \textbf{10.52} & \textbf{6.11} & \textbf{0.787} \\
\bottomrule
\end{tabular}
\caption{Absolute Numerical Prediction on Infant Mortality ($N=100$ for baselines).}
\label{tab:absolute_pred}
\end{table}

\section{More Visualizations}

\begin{figure*}[htb!] 
    \centering 
    \includegraphics[width=0.9\textwidth]{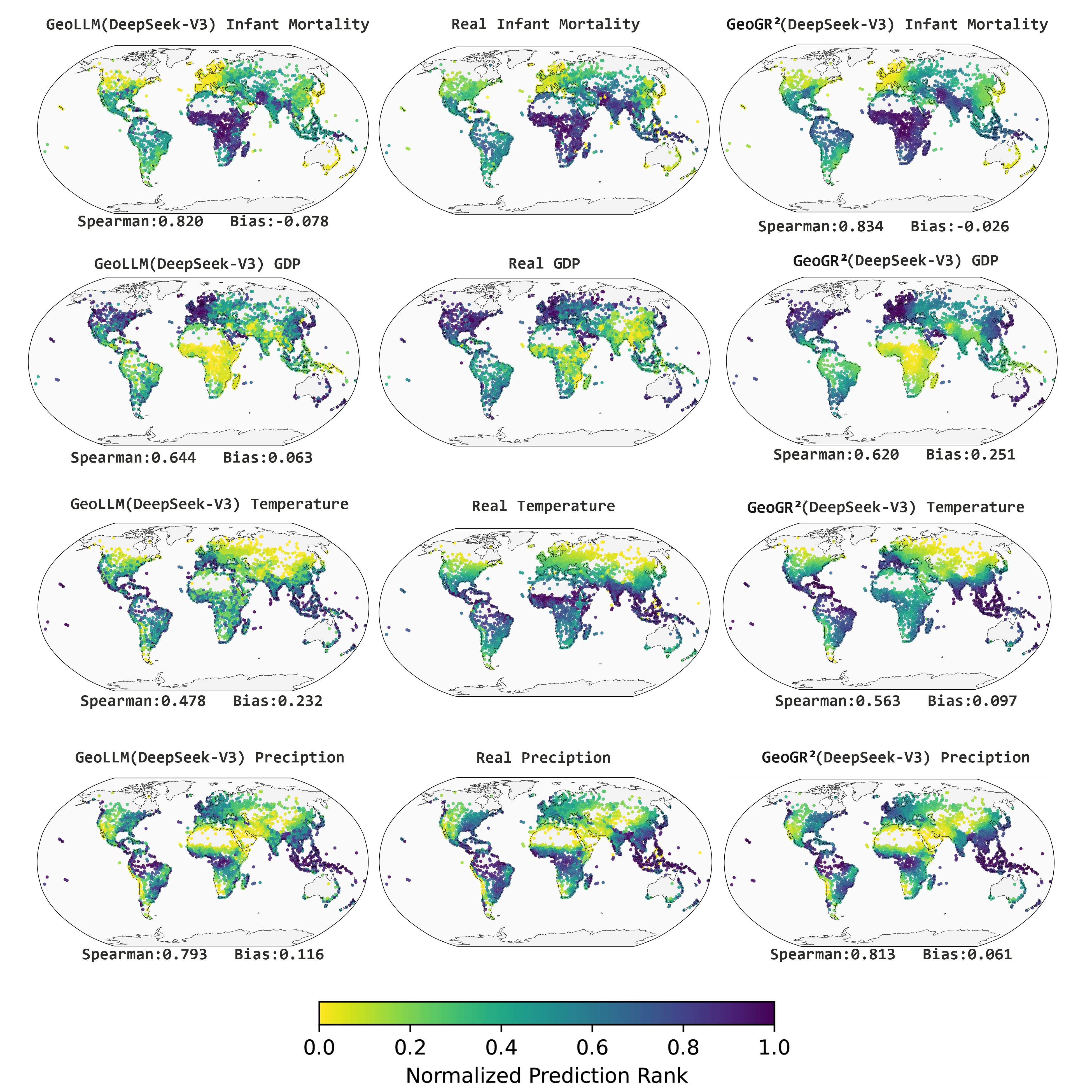} 
    \caption{Spatial prediction rank maps using DeepSeek-V3. \textbf{Left}: Predictions from the base GeoLLM.
    \textbf{Middle}: Ground-truth. \textbf{Right}:Predictions from GeoGR²-enhanced model. } 
    \label{app:deepseek}
\end{figure*}

\begin{figure*}[htb!] 
    \centering 
    \includegraphics[width=0.9\textwidth]{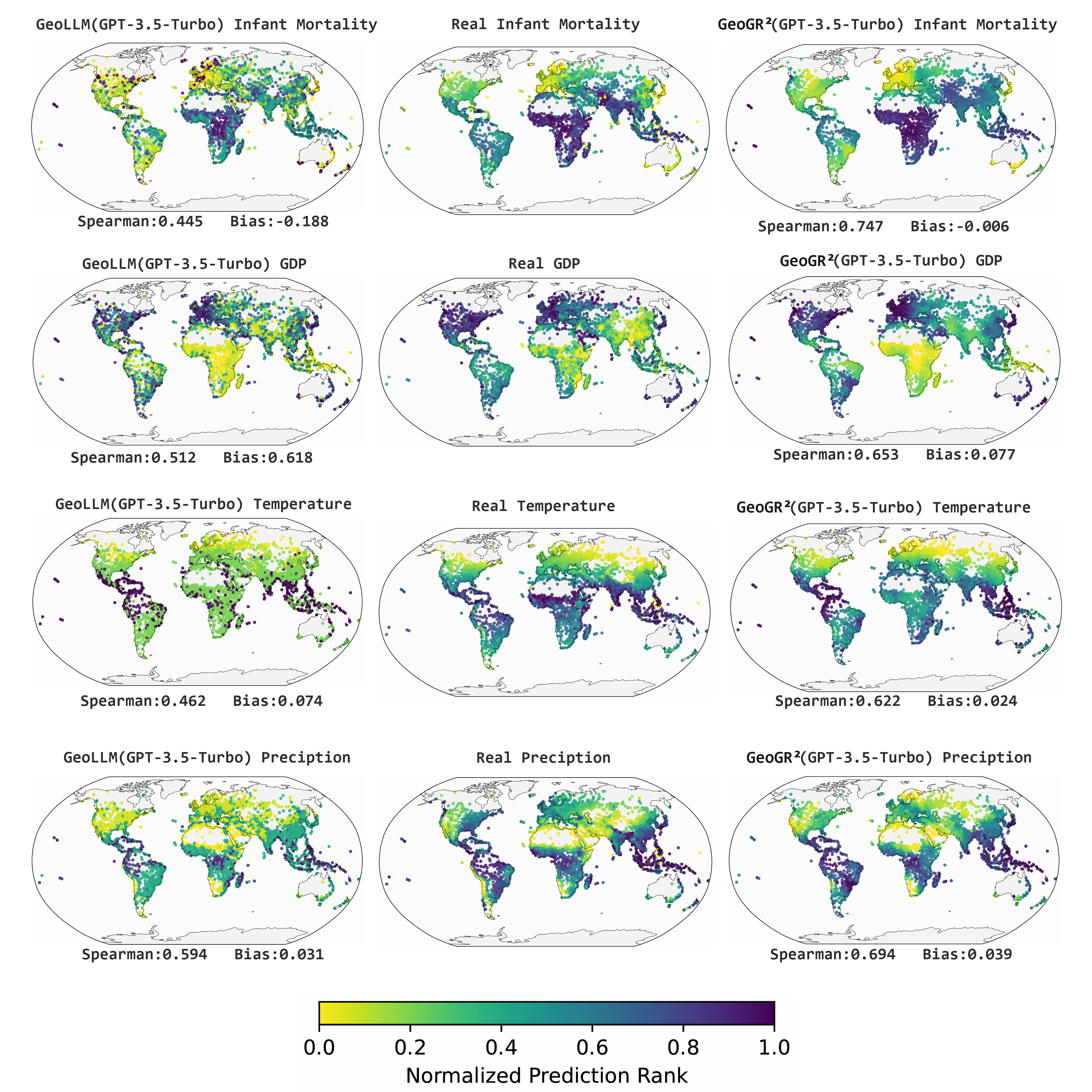} 
    \caption{Spatial prediction rank maps using GPT-3.5-Turbo. \textbf{Left}: Predictions from the base GeoLLM.
    \textbf{Middle}: Ground-truth. \textbf{Right}: Predictions from GeoGR²-enhanced model.} 
    \label{app:gpt35_vis}
\end{figure*}

\begin{figure*}[htb!] 
    \centering 
    \includegraphics[width=0.9\textwidth]{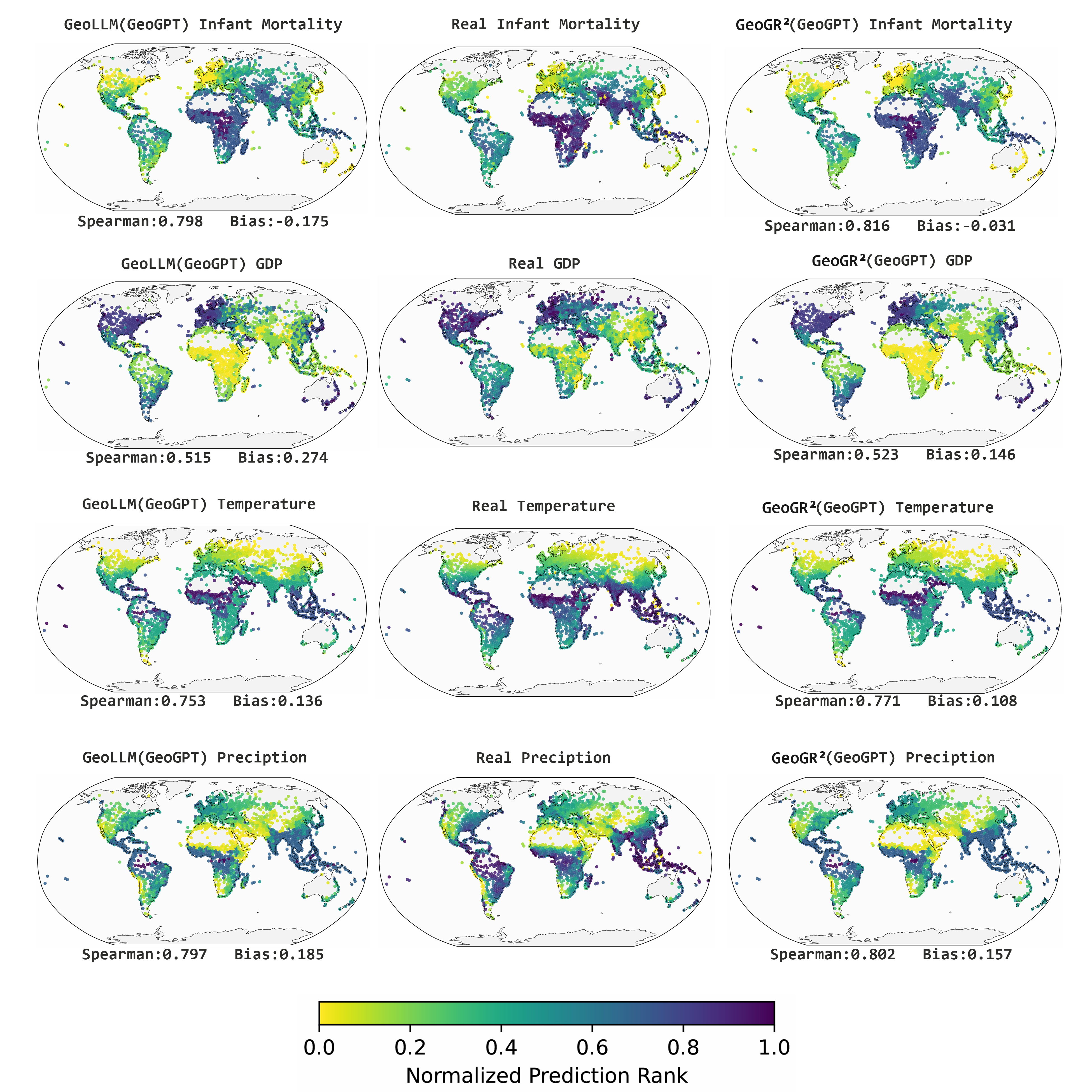} 
    \caption{Spatial prediction rank maps using GeoGPT. \textbf{Left}: Predictions from the base GeoLLM.
    \textbf{Middle}: Ground-truth. \textbf{Right}: Predictions from GeoGR²-enhanced model.} 
    \label{app:geogpt}
\end{figure*}

We provide visualizations of all models in the \cref{app:deepseek},  \cref{app:gpt35_vis} and \cref{app:geogpt}, and we find that GeoGR² consistently enhances the overall performance of the models. In some tasks, such as those involving GPT-3.5-Turbo, GeoGR² significantly improves the model's effectiveness. However, we observe that large language models still exhibit tendencies to systematically overestimate or underestimate certain regions; for instance, all models tend to overestimate China's GDP, and although GeoGR² mitigates this to some extent, the overall bias remains upward.

\section{More Ablation Studies on Point Selection Strategy}

In the main text, we provided an ablation study for GPT-3.5-turbo. We also included results from applying the Point Selection Strategy to three additional models: DeepSeek V3, GPT-4o-mini, and GeoGPT. The findings in \cref{tab:pointDeepSeek}, \cref{tab:pointGPT4Omini}, and \cref{tab:pointGEOgpt} show that the impact of the Point Selection Strategy on these models is consistent with its effect on GPT-3.5-turbo, with comparable overall performance across all models.

\begin{table}[h]
\centering
\caption{Ablation Study on Point Selection Strategy for DeepSeek-V3}
\label{tab:pointDeepSeek}
\scriptsize  
\begin{tabular}{@{}llcccc@{}}
\toprule
\multicolumn{2}{c}{Method Variant} & \multicolumn{4}{c}{Tasks} \\
\cmidrule(lr){1-2} \cmidrule(lr){3-6}
Method & Metric & \shortstack{Infant mort.} & GDP & \shortstack{Temp.} & \shortstack{Precip.} \\
\midrule
\multirow{2}{*}{\shortstack[l]{GeoGR² w/o\\ ${N}_t^{\text{prox}}$}} 
    & Spearman &0.822 &0.632 &0.504 &0.809 \\
    & Bias     &-0.057 &0.192 &0.106 &0.097 \\
\addlinespace[0.2em]
\multirow{2}{*}{\shortstack[l]{GeoGR² w/o\\ ${N}_t^{\text{sem}}$}} 
    & Spearman &0.830 &0.641 &0.560 &0.815 \\
    & Bias     &-0.021 &0.057 &0.089 &0.044 \\
\addlinespace[0.2em]
\multirow{2}{*}{GeoGR²} 
    & Spearman &\textbf{0.834} &\textbf{0.644} &\textbf{0.563} &\textbf{0.813} \\
    & Bias    &-0.026 &0.063 &0.097 &0.061 \\
\bottomrule
\end{tabular}
\end{table}

\begin{table}[h]
\centering
\caption{Ablation Study on Topology Construction Strategy for GPT-4o-mini}
\label{tab:pointGPT4Omini}
\scriptsize  
\begin{tabular}{@{}llcccc@{}}
\toprule
\multicolumn{2}{c}{Method Variant} & \multicolumn{4}{c}{Tasks} \\
\cmidrule(lr){1-2} \cmidrule(lr){3-6}
Method & Metric & \shortstack{Infant mort.} & GDP & \shortstack{Temp.} & \shortstack{Precip.} \\
\midrule
\multirow{2}{*}{\shortstack[l]{GeoGR² w/o\\ ${N}_t^{\text{prox}}$}} 
    & Spearman &0.730 &0.581 &0.504 &0.631 \\
    & Bias     &-0.052 &0.192 &0.102 &0.097 \\
\addlinespace[0.2em]
\multirow{2}{*}{\shortstack[l]{GeoGR² w/o\\ ${N}_t^{\text{sem}}$}} 
    & Spearman &0.744 &0.577 &0.510 &0.626 \\
    & Bias     &-0.009 &0.057 &0.047 &0.044 \\
\addlinespace[0.2em]
\multirow{2}{*}{GeoGR²} 
    & Spearman &\textbf{0.762} &\textbf{0.585} &\textbf{0.517} &\textbf{0.634} \\
    & Bias    &-0.011 &0.087 &0.085 &0.011 \\
\bottomrule
\end{tabular}
\end{table}

\begin{table}[h]
\centering
\caption{Ablation Study on Topology Construction Strategy for GeoGPT}
\label{tab:pointGEOgpt}
\scriptsize  
\begin{tabular}{@{}llcccc@{}}
\toprule
\multicolumn{2}{c}{Method Variant} & \multicolumn{4}{c}{Tasks} \\
\cmidrule(lr){1-2} \cmidrule(lr){3-6}
Method & Metric & \shortstack{Infant mort.} & GDP & \shortstack{Temp.} & \shortstack{Precip.} \\
\midrule
\multirow{2}{*}{\shortstack[l]{GeoGR² w/o\\ ${N}_t^{\text{prox}}$}} 
    & Spearman &0.801 &0.517 &0.758 &0.799 \\
    & Bias     &-0.022 &0.127 &0.112 &0.160 \\
\addlinespace[0.2em]
\multirow{2}{*}{\shortstack[l]{GeoGR² w/o\\ ${N}_t^{\text{sem}}$}} 
    & Spearman &0.811 &0.525 &0.769 &0.807 \\
    & Bias     &-0.011 &0.088 &0.099 &0.144 \\
\addlinespace[0.2em]
\multirow{2}{*}{GeoGR²} 
    & Spearman &\textbf{0.816} &0.523 &\textbf{0.771} &0.802 \\
    & Bias    &{-0.031} &{0.146} &{0.108} &{0.157} \\
\bottomrule
\end{tabular}
\end{table}


\section{More Ablation Studies on External Variable}

We use the 19 bioclimatic variables (bio1–bio19) from WorldClim v2.1\footnote{\url{https://www.worldclim.org/}} as spatially explicit covariates. Derived from long-term monthly temperature and precipitation records, these indices capture both mean climatic conditions and seasonal extremes. \cref{tab:bio_vars} provides concise ecological definitions.

\begin{table*}[h!]
\centering
\caption{WorldClim bioclimatic variables used in this study.}
\label{tab:bio_vars}
\small
\begin{tabular}{cll}
\toprule
ID & Variable & Ecological relevance \\
\midrule
\multicolumn{3}{l}{\textbf{Temperature-related (energy)}}\\
bio1 & Annual Mean Temperature & Baseline thermal energy \\
bio2 & Mean Diurnal Range & Daily temperature stability \\
bio3 & Isothermality & Shape of thermal variability \\
bio4 & Temperature Seasonality & Intensity of thermal pulses \\
bio5 & Max Temperature of Warmest Month & Acute heat stress \\
bio6 & Min Temperature of Coldest Month & Acute cold stress \\
bio7 & Temperature Annual Range & Climatic buffering capacity \\
bio8 & Mean Temp.\ of Wettest Quarter & Thermal conditions under high moisture \\
bio9 & Mean Temp.\ of Driest Quarter & Thermal conditions under low moisture \\
bio10 & Mean Temp.\ of Warmest Quarter & Prolonged heat exposure \\
bio11 & Mean Temp.\ of Coldest Quarter & Prolonged cold exposure \\
\midrule
\multicolumn{3}{l}{\textbf{Precipitation-related (water)}}\\
bio12 & Annual Precipitation & Total water input \\
bio13 & Precipitation of Wettest Month & Moisture extremes \\
bio14 & Precipitation of Driest Month & Moisture extremes \\
bio15 & Precipitation Seasonality & Drought risk \\
bio16 & Precipitation of Wettest Quarter & Seasonal water allocation \\
bio17 & Precipitation of Driest Quarter & Seasonal water allocation \\
bio18 & Precipitation of Warmest Quarter & Temperature–precipitation coupling \\
bio19 & Precipitation of Coldest Quarter & Temperature–precipitation coupling \\
\bottomrule
\end{tabular}
\end{table*}

In the main text, we provided an ablation study for GPT-3.5-turbo. We also included results from applying the Variable Selection Strategy to three additional models: DeepSeek V3, GPT-4o-mini, and GeoGPT. The findings in \cref{tab:bioDeepSeek}, ~\ref{tab:bioGPT4Omini}, and ~\ref{tab:bioGEOgpt} show that the impact of the Variable Selection Strategy on these models is consistent with its effect on GPT-3.5-turbo, with comparable overall performance across all models.

\begin{table}[h]
\centering
\caption{Ablation Study on External Variables for DeepSeek-V3}
\label{tab:bioDeepSeek}
\scriptsize  
\begin{tabular}{@{}llcccc@{}}
\toprule
\multicolumn{2}{c}{Method Variant} & \multicolumn{4}{c}{Tasks} \\
\cmidrule(lr){1-2} \cmidrule(lr){3-6}
Method & Metric & \shortstack{Infant mort.} & GDP & \shortstack{Temp.} & \shortstack{Precip.} \\
\midrule
\multirow{2}{*}{\shortstack[l]{GeoGR² w/o\\ external variables}} 
  & Spearman &0.825 &0.628 &0.579 &0.799 \\
    & Bias     &-0.036 &0.093 &0.101 &0.086 \\
\addlinespace[0.2em]
\multirow{2}{*}{GeoGR²} 
    & Spearman &\textbf{0.834} &\textbf{0.644} &\textbf{0.563} &\textbf{0.813} \\
    & Bias    &\textbf{-0.026} &\textbf{0.063} &\textbf{0.097} &\textbf{0.061} \\
\bottomrule
\end{tabular}
\end{table}

\begin{table}[h]
\centering
\caption{Ablation Study on External Variables for GPT-4o-mini}
\label{tab:bioGPT4Omini}
\scriptsize  
\begin{tabular}{@{}llcccc@{}}
\toprule
\multicolumn{2}{c}{Method Variant} & \multicolumn{4}{c}{Tasks} \\
\cmidrule(lr){1-2} \cmidrule(lr){3-6}
Method & Metric & \shortstack{Infant mort.} & GDP & \shortstack{Temp.} & \shortstack{Precip.} \\
\midrule
\multirow{2}{*}{\shortstack[l]{GeoGR² w/o\\ external variables}} 
  & Spearman &0.710 &0.554 &0.524 &0.601 \\
    & Bias     &-0.033 &0.091 &0.062 &0.059 \\
\addlinespace[0.2em]
\multirow{2}{*}{GeoGR²} 
    & Spearman &\textbf{0.762} &\textbf{0.585} &\textbf{0.517} &\textbf{0.634} \\
    & Bias    &\textbf{-0.011} &\textbf{0.087} &{0.085} &\textbf{0.011} \\
\bottomrule
\end{tabular}
\end{table}

\begin{table}[h]
\centering
\caption{Ablation Study on External Variables for GeoGPT}
\label{tab:bioGEOgpt}
\scriptsize  
\begin{tabular}{@{}llcccc@{}}
\toprule
\multicolumn{2}{c}{Method Variant} & \multicolumn{4}{c}{Tasks} \\
\cmidrule(lr){1-2} \cmidrule(lr){3-6}
Method & Metric & \shortstack{Infant mort.} & GDP & \shortstack{Temp.} & \shortstack{Precip.} \\
\midrule
\multirow{2}{*}{\shortstack[l]{GeoGR² w/o\\ external variables}} 
  & Spearman &0.801 &0.517 &0.781 &0.800 \\
    & Bias     &-0.043 &0.095 &0.134 &0.116 \\
\addlinespace[0.2em]
\multirow{2}{*}{GeoGR²} 
    & Spearman &\textbf{0.816} &\textbf{0.523} &{0.771} &\textbf{0.802} \\
    & Bias    &\textbf{-0.031} &{0.146} &\textbf{0.108} &{0.157} \\
\bottomrule
\end{tabular}
\end{table}
\section{More Ablation Studies on Iteration Rounds Experiment}
\begin{figure*}[htb!] 
    \centering 
    \includegraphics[width=0.9\textwidth]{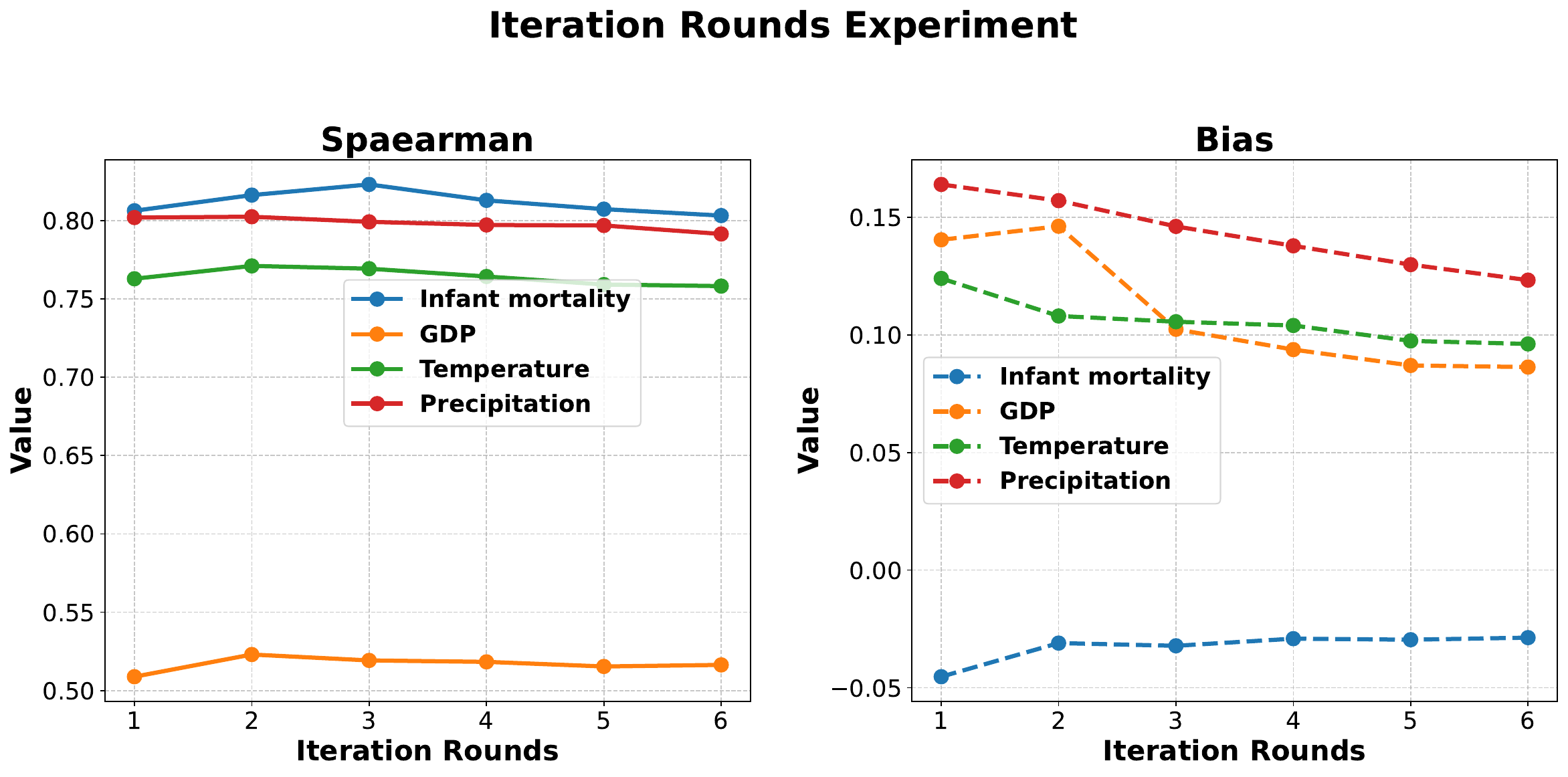} 
    \caption{Iteration Rounds Experiment(GeoGR²-enhanced GeoGPT). } 
    \label{iteration:GeoGPT}
\end{figure*}

\begin{figure*}[htb!] 
    \centering 
    \includegraphics[width=0.9\textwidth]{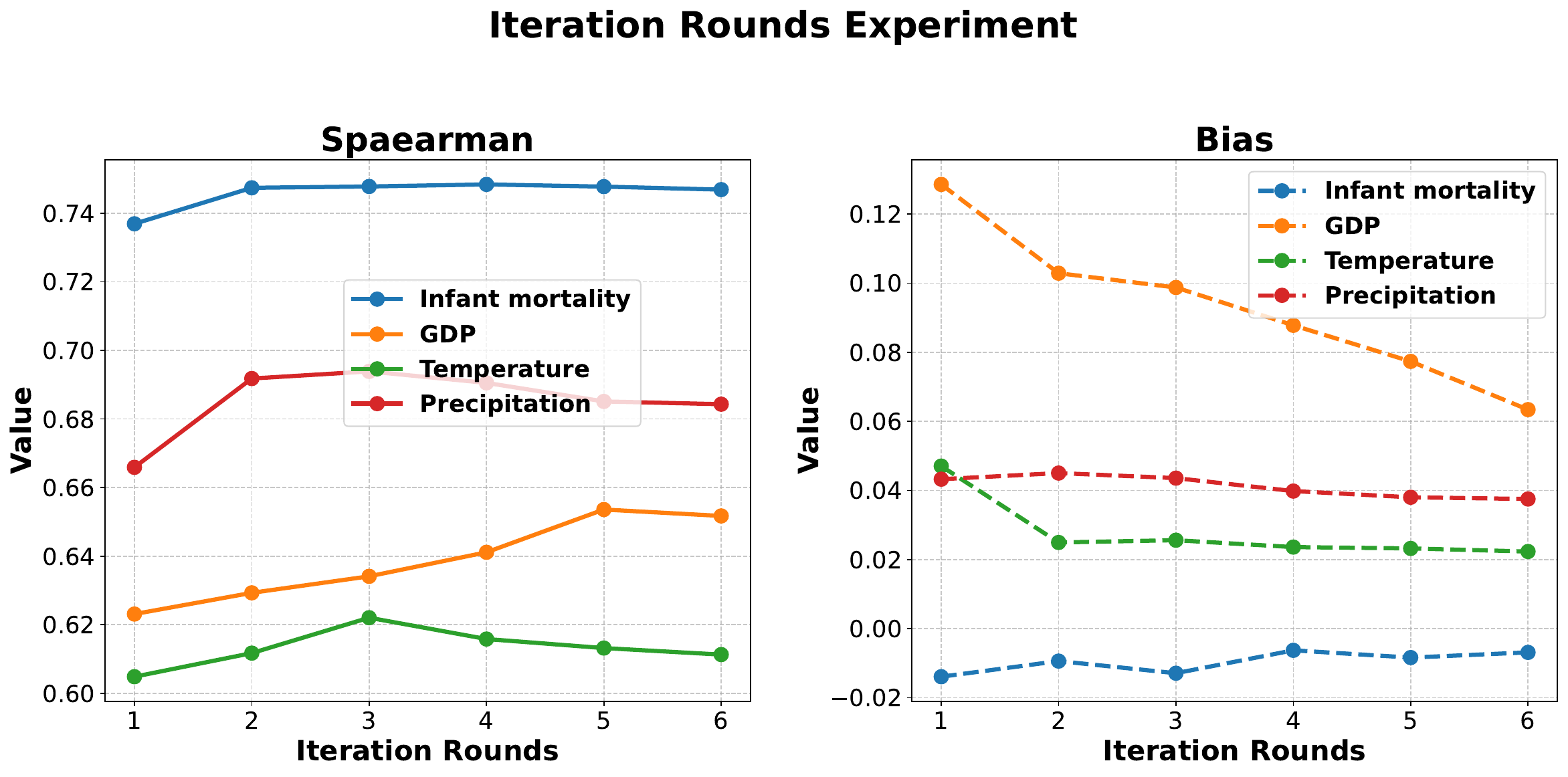} 
    \caption{Iteration Rounds Experiment(GeoGR²-enhanced GPT-3.5-Turbo). } 
    \label{iteration:GPT-3.5}
\end{figure*}

\begin{figure*}[htb!] 
    \centering 
    \includegraphics[width=0.9\textwidth]{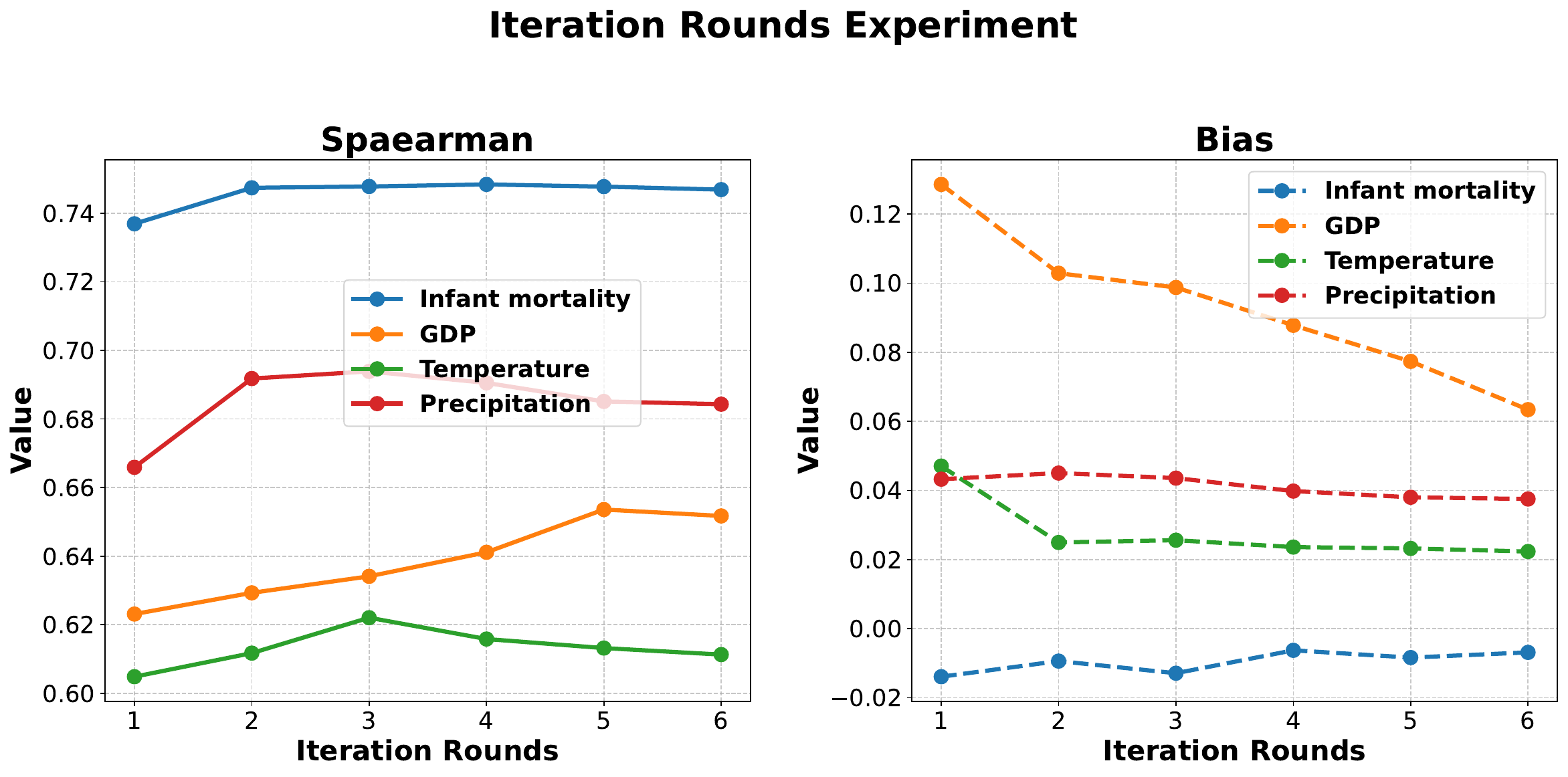} 
    \caption{Iteration Rounds Experiment(GeoGR²-enhanced GPT-4o-mini). } 
    \label{iteration:GPT-4o-mini}
\end{figure*}

we also provide iteration rounds experiment results of three additional models: GeoGPT, GPT-4o-mini, and GPT-3.5-Turbo.As shown in \cref{iteration:GeoGPT}, \ref{iteration:GPT-3.5} and \ref{iteration:GPT-4o-mini}, the Spearman values of all models reached their maximum after several iterations and then began to decline, while the bias scores continued to decrease with further iterations.

\begin{figure*}[htb!] 
    \centering 
    \includegraphics[width=0.9\textwidth]{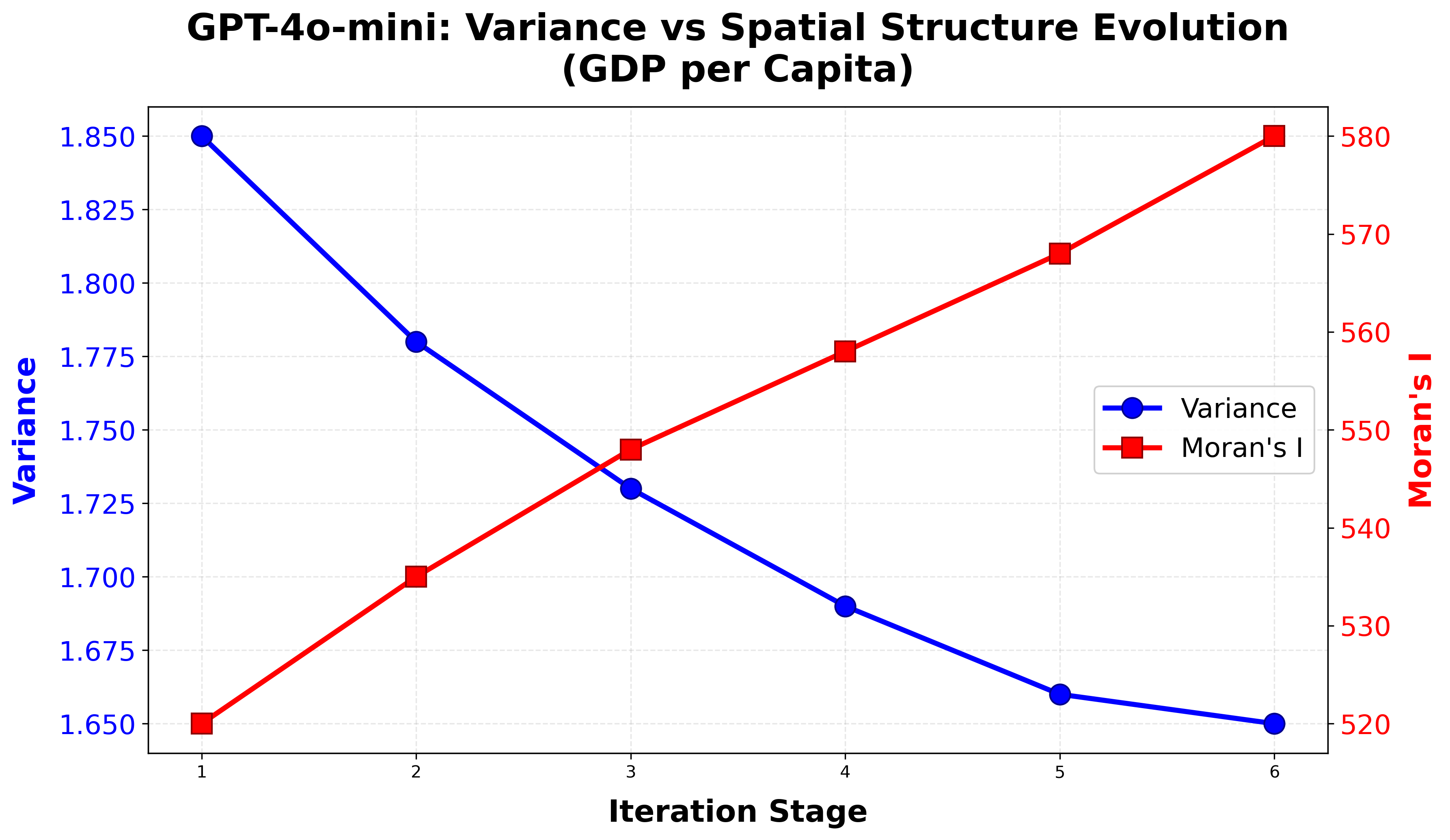} 
    \caption{patial heterogeneity and convergence metrics
across refinement rounds (GDP task, GPT-4o-
mini). } 
    \label{moran:GPT-4o-mini_gdp}
\end{figure*}

\begin{figure*}[htb!] 
    \centering 
    \includegraphics[width=0.9\textwidth]{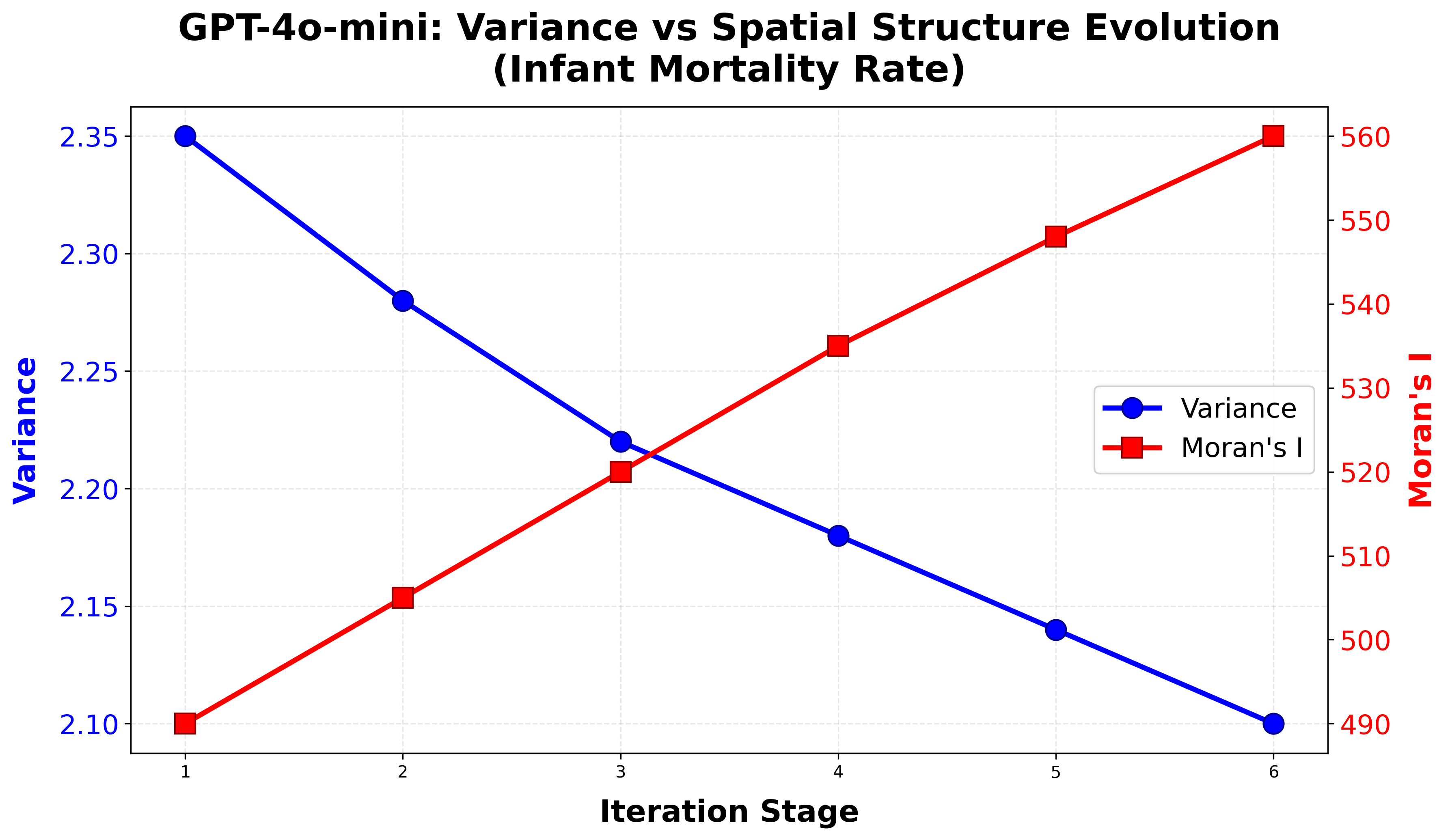} 
    \caption{patial heterogeneity and convergence metrics
across refinement rounds (Infant Mortality task, GPT-4o-
mini). } 
    \label{moran:GPT-4o-mini_infant}
\end{figure*}

\begin{figure*}[htb!] 
    \centering 
    \includegraphics[width=0.9\textwidth]{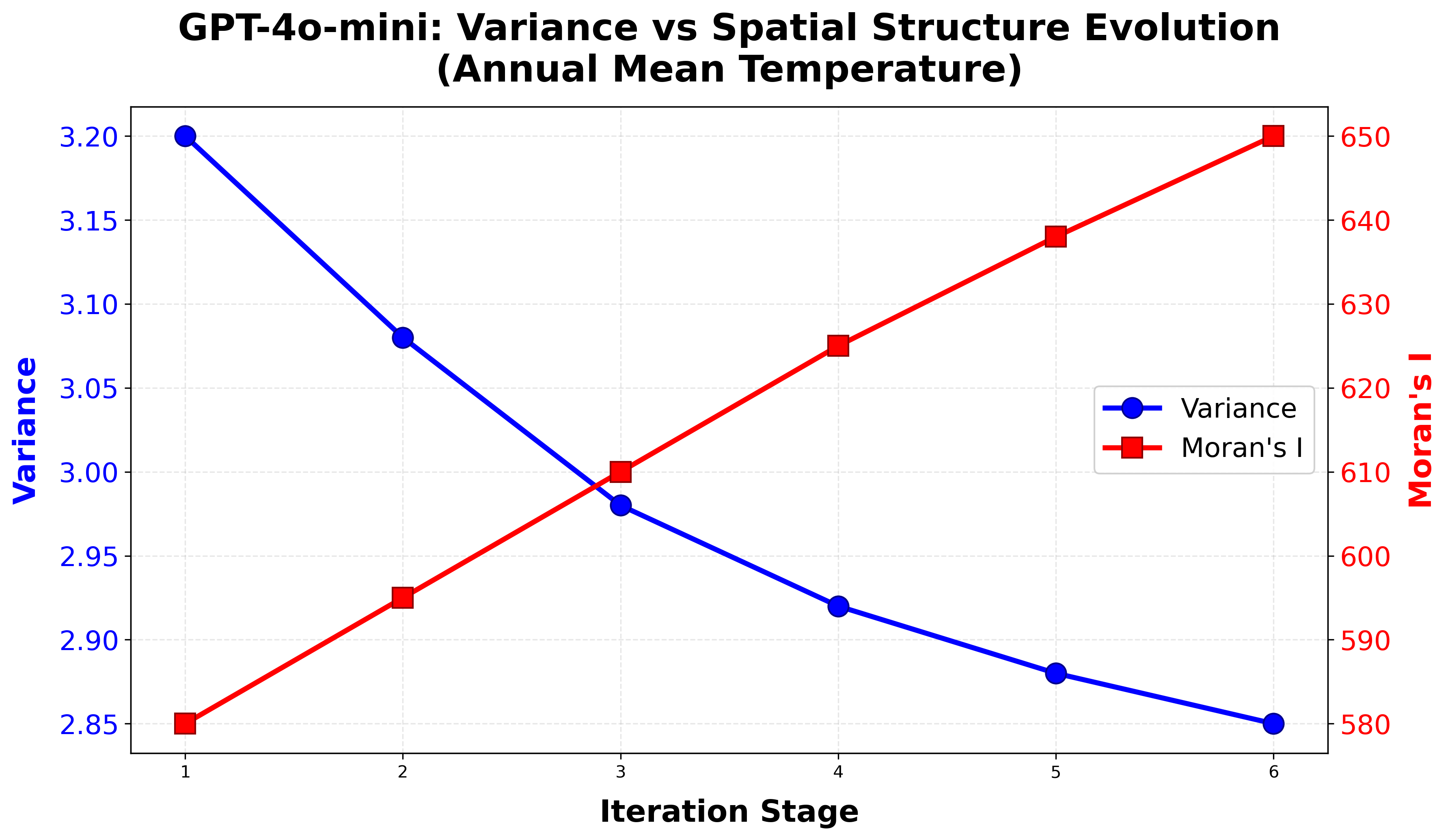} 
    \caption{patial heterogeneity and convergence metrics
across refinement rounds (Temperature task, GPT-4o-
mini). } 
    \label{moran:GPT-4o-mini_temperature}
\end{figure*}

\end{document}